\documentclass[11pt]{article}

\usepackage[final]{acl}

\usepackage{times}
\usepackage{latexsym}
\usepackage{ragged2e}

\usepackage[T1]{fontenc}

\usepackage[utf8]{inputenc}

\usepackage{microtype}

\usepackage{inconsolata}

\usepackage{graphicx}

\usepackage{enumitem}
\usepackage{amsmath}
\usepackage{booktabs}
\usepackage{subcaption}
\usepackage{multirow}
\usepackage{CJKutf8}

\newcommand{\framework}{Re-RIGHT}

\title{Right at My Level: A Unified Multilingual Framework for Proficiency-Aware Text Simplification}

\author{
 \textbf{Jinhong Jeong\textsuperscript{1}} \quad
 \textbf{Junghun Park\textsuperscript{2}} \quad
 \textbf{Youngjae Yu\textsuperscript{2}}
\\
 \textsuperscript{1}Yonsei University \quad
 \textsuperscript{2}Seoul National University
\\
\\
\texttt{jjhsnail0822@yonsei.ac.kr}
}

\begin{document}
\maketitle

\begin{abstract}

Text simplification supports second language (L2) learning by providing comprehensible input, consistent with the Input Hypothesis. However, constructing personalized parallel corpora is costly, while existing large language model (LLM)-based readability control methods rely on pre-labeled sentence corpora and primarily target English. We propose \textbf{\framework}, a unified reinforcement learning framework for adaptive multilingual text simplification without parallel corpus supervision. We first show that prompting-based lexical simplification at target proficiency levels (CEFR, JLPT, TOPIK, and HSK) performs poorly at easier levels and for non-English languages, even with state-of-the-art LLMs such as GPT-5.2 and Gemini 2.5. To address this, we collect 43K vocabulary-level data across four languages (English, Japanese, Korean, and Chinese) and train a compact 4B policy model using \framework, which integrates three reward modules: vocabulary coverage, semantic preservation, and coherence. Compared to the stronger LLM baselines, \framework\ achieves higher lexical coverage at target proficiency levels while maintaining original meaning and fluency.

\end{abstract}

\section{Introduction}

\begin{figure*}[tb]
    \centering
    \includegraphics[width=\textwidth]{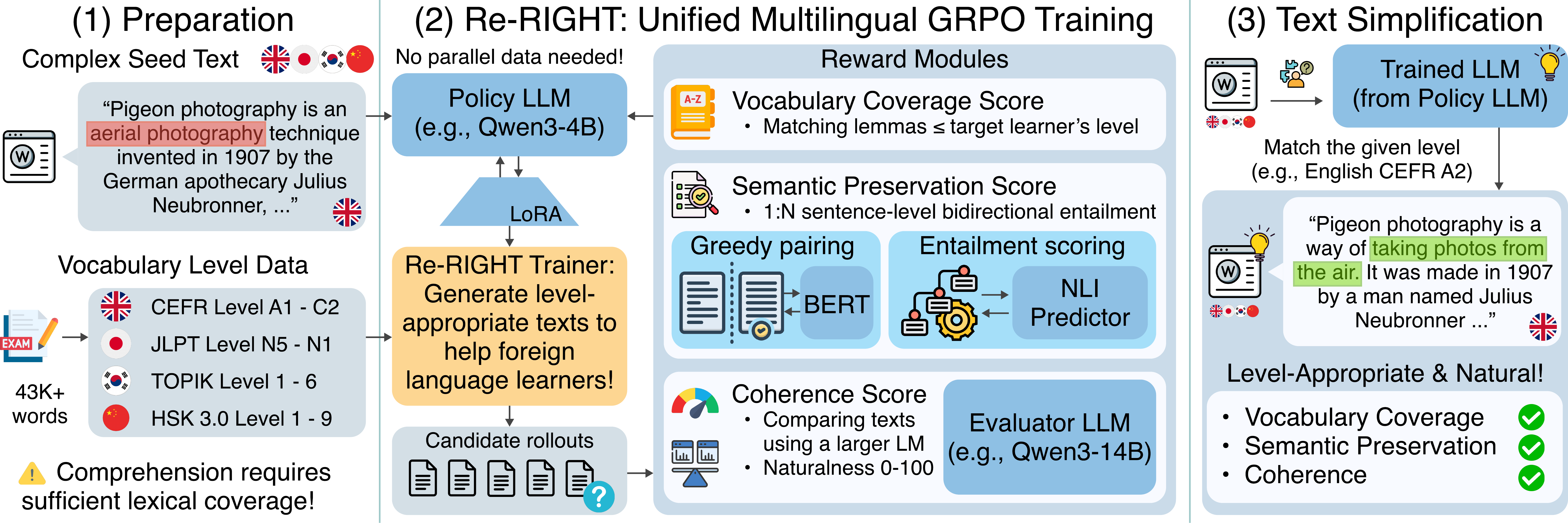}
    \caption{We propose \textbf{\framework}, a unified multilingual GRPO training framework for text simplification at learners' proficiency levels (\S\ref{sec:rl_based_improvement_experiments}) without relying on parallel training corpora, overcoming the limitation of the prompting-based approach (\S\ref{sec:can_current_llms_generate_level_appropriate_texts}) that fails to meet the target vocabulary coverage. (1) At the preparation phase, we collect ``Featured Articles'' in Wikipedia as a training seed dataset, while constructing 43,786 vocabulary level data from official proficiency standards across four languages. (2) We then train a 4B policy model with three reward modules: vocabulary coverage, semantic preservation, and coherence. (3) The trained policy model enables adaptive multilingual text simplification with higher vocabulary coverage than GPT-5.2 and Gemini 2.5, generating level-appropriate, semantically preserved, and coherent outputs.}
    \label{fig:figure_1}
\end{figure*}

Second language (L2) acquisition is most effective when language learners are exposed to sufficient ``comprehensible input'' at levels between their current proficiency ($i$) and the marginally higher level ($i + 1$), according to the Input Hypothesis~\citep{krashen1981second}. Linguistic research has further reported that L2 learners need to know 95--98\% of the words in a given text to achieve fluent reading comprehension~\citep{Hu2000, schmitt2011percentage}. Accordingly, adequate vocabulary coverage plays a crucial role in providing level-appropriate language input. In real educational environments, however, continuously supplying level-specific reading materials requires considerable effort from professionals, making it impractical in terms of cost and time.

Text simplification addresses this problem by controlling lexical complexity to match a target reader's vocabulary knowledge while preserving original meaning~\citep{10.1145/3442695}. Despite substantial advances in large language models (LLMs) and their strong zero-shot capabilities~\citep{wei2021finetuned}, they still fall short of reliably generating text that is precisely tailored to a specific proficiency levels~\citep{barayan2025analysing}.

To overcome these limits, recent studies have explored reinforcement learning (RL)-based approaches to improve controllability in text simplification~\citep{zhang-lapata-2017-sentence, malik-etal-2024-tarzan, li-etal-2025-aligning}. However, these methods require pre-labeled, level-specific sentences and are predominantly restricted to English, thereby offering limited applicability in multilingual environments.

In this context, we introduce \textbf{\framework} 
(\textbf{Re}inforcement learning for \textbf{R}eadability \textbf{I}mprove\-ment via \textbf{G}eneration from \textbf{H}ard \textbf{T}exts),
an adaptive and unified multilingual text simplification framework for vocabulary learning that does not rely on predefined sentence corpora. Our framework supports four languages (English, Japanese, Korean, and Chinese) from different language families and writing systems, yet sharing similar practical learning needs~\citep{yang2003motivational}.

To systematically control lexical proficiency, we construct a vocabulary level dataset based on the standardized proficiency scales (CEFR, JLPT, TOPIK, and HSK 3.0), and quantitatively define a vocabulary coverage score as the proportion of level-appropriate vocabulary within the content words of a given text. Using this criterion, we first examine the ability of state-of-the-art LLMs like GPT-5.2 and Gemini 2.5 to generate texts that match target proficiency levels. Our findings reveal that LLMs struggle particularly with easy proficiency levels, where this performance gap becomes even larger in non-English languages. We address this challenge by training a small, unified policy model using \framework. Our main contributions are summarized as follows:

\begin{enumerate}
    \item We propose \framework, an adaptive, multilingual text simplification framework that supports language learning through level-appropriate vocabulary without relying on pre-labeled sentences, as illustrated in Figure~\ref{fig:figure_1}.
    \item We provide a comprehensive multilingual benchmark for LLM-based text simplification using 43K+ vocabulary level data, demonstrating that even state-of-the-art LLMs struggle to reliably generate texts at easy proficiency levels and in non-English settings.
    \item We introduce three novel reward modules in \framework, integrating (1) vocabulary coverage grounded in standardized proficiency scales, (2) semantic preservation based on a 1:N sentence-level bidirectional entailment score, and (3) coherence reward through comparative prompting.
\end{enumerate}

Experimental results demonstrate that a 4B policy model with \framework\ significantly increases the proportion of level-appropriate vocabulary while preserving original meaning and fluency. Our study addresses the fundamental limitation of current LLMs in generating level-appropriate texts, providing a generalizable foundation toward practical L2 learning support across multiple languages.

\section{Related Work}

\subsection{Controlled Vocabulary in Applied Linguistics}

The idea of controlling vocabulary for language learning can be traced back to the ``Basic English'' developed by \citet{ogden1930basic}, who suggested 850 core words for simplified English. Since then, contemporary applied linguists have consistently shown that achieving sufficient (mainly 95--98\%) lexical coverage is essential for effective comprehension of L2 input~\citep{laufer1989percentage, Hu2000, schmitt2011percentage, Durbahn_Rodgers_Macis_Peters_2024}, implying that learners' lexical knowledge plays a pivotal role in providing comprehensible input based on the Input Hypothesis~\citep{krashen1981second, webb2021research}. Studies have also reported strong statistical correlations between L2 learners' vocabulary knowledge and their reading comprehension performance~\citep{zhang2012vocabulary, jeon2022l2, zhang2022relationship}. These findings have been utilized in the level classification of authoritative language proficiency standards such as CEFR and learner's vocabulary lists~\citep{Milton2010TheDO, 10.1093/applin/amt018, capel2015english}, clearly demonstrating the impact of appropriate vocabulary control in language education.

\subsection{Text Simplification via Language Models}

Researchers have widely investigated language model based text simplification~\citep{north2025deep}, constructing readability-controlled datasets at various difficulty levels for English~\citep{kogan2025acecefrdatasetautomated} and non-English languages~\citep{naous-etal-2024-readme, imperial2025universalcefrenablingopenmultilingual, anschütz2025german4alldatasetmodel}. Several studies have also introduced LLM-based benchmarks~\citep{ryan-etal-2023-revisiting, maddela-etal-2023-lens, kew-etal-2023-bless} or agentic simplification framework~\citep{mo-hu-2024-expertease}. Since applying simple prompt engineering approach has been viewed as insufficient for the task~\citep{barayan2025analysing}, controlled decoding strategies~\citep{kew-ebling-2022-target, zetsu-etal-2022-lexically} and RL-based LLM training~\citep{yanamoto-etal-2022-controllable, malik-etal-2024-tarzan} have gained prominence.

Specifically, \citet{li-etal-2025-aligning} adopts PPO algorithm~\citep{schulman2017proximalpolicyoptimizationalgorithms} to train text simplification models that require data labeled at each CEFR level instead of explicit parallel corpora, with the purpose of generating level-appropriate texts for English as a Second Language (ESL) learners. Our work expands this research by utilizing GRPO algorithm~\citep{shao2024deepseekmathpushinglimitsmathematical} with linguistically designed reward modules to eliminate the necessity for predetermined sentence complexity labels, thereby facilitating unified adaptability in multilingual readability control at specific proficiency levels. Furthermore, our framework ensures naturalness in simplified outputs by measuring coherence through prompting multilingual LLMs, unlike \citet{siddharthan-2003-preserving}, \citet{vasquez-rodriguez-etal-2023-document}, and \citet{vasquezrodriguez2024simpleenoughdocumentleveltext}, which rely on algorithmic approaches or fine-tuned language models.

\section{Datasets}
\label{sec:datasets}

\subsection{Multilingual Vocabulary Level Data}
\label{sec:multilingual_vocabulary_level_data}

We collect vocabulary lists for four languages (English, Japanese, Korean, and Chinese), according to each language's authoritative proficiency levels. We used Common European Framework of Reference (CEFR)~\citep{council2001common} for English, Japanese Language Proficiency Test (JLPT)~\citep{jlpt_official_website} for Japanese, Test of Proficiency in Korean (TOPIK)~\citep{topik_official_website} for Korean, and Hanyu Shuiping Kaoshi (HSK) 3.0~\citep{chinesetest_official_website}\footnote{Levels of HSK7, HSK8, and HSK9 are integrated into the same HSK7-9 level in accordance with the official vocabulary list's formulation.} for Chinese as our vocabulary level criteria.

For vocabulary level data, English data are based on the English Vocabulary Profile~\citep{capel2015english}, while Korean and Chinese data are derived from official vocabulary lists published by national institutions~\citep{nakl_2017_korean_standard_curriculum, moe_china_2021_education_notice}. Since no official level-specific vocabulary list exists for Japanese, we utilize a widely adopted unofficial online resource.\footnote{\url{https://www.tanos.co.uk/}}

After collecting the data, words are lemmatized using the spaCy~\citep{Honnibal_spaCy_Industrial-strength_Natural_2020} library.\footnote{Except Japanese, where lemmatization for standalone words yields low performance, so source lemmas are used.} In cases where multiple levels are assigned to the same lemma, the lowest level is used as the basis, resulting in a total of 43,786 lemmas. The summary of the proficiency level data is shown in Table~\ref{tab:vocab_level_list} in the Appendix~\ref{sec:details_on_the_datasets}.

\subsection{Wikipedia's Featured Article Data}
\label{sec:non_parallel_training_data_construction}

In contrast to prior studies, our framework does not require level-annotated training data. Instead, we refine the ``Featured Articles'' in Wikipedia\footnote{\url{https://www.wikipedia.org/}} as a seed dataset, which contains high-quality and sufficiently complex texts for each language.

We extract a total of 8,057 articles from Wikipedia across four languages. Considering the training memory, we chunk each article into units of up to 512 policy model tokens, while preserving paragraph boundaries to maintain contextual coherence. We then generate multiple proficiency-level variations for each text and uniformly sample across languages. Consequently, this process results in 69,220 data chunks, of which we use 10\% as the test set and the remainder as the training set.

\subsection{Additional Test Set: Parallel Global Voices}
\label{sec:additional_test_set}

In addition to the Wikipedia Featured Article test set,\footnote{This data is used as a main training and test set throughout this work unless otherwise specified.} we examine the generalizability of our framework by constructing a test set from another domain, namely news data. We use Parallel Global Voices dataset~\citep{PROKOPIDIS16.778}, which is preprocessed in the same manner as in \S\ref{sec:non_parallel_training_data_construction}. We uniformly sample 300 level-duplicated news chunks for each language, building a test set with a total of 1,200 documents. Details on the dataset construction in \S\ref{sec:datasets} are provided in the Appendix~\ref{sec:details_on_the_datasets}.

\begin{figure*}[tb]
    \centering
    \includegraphics[width=\textwidth]{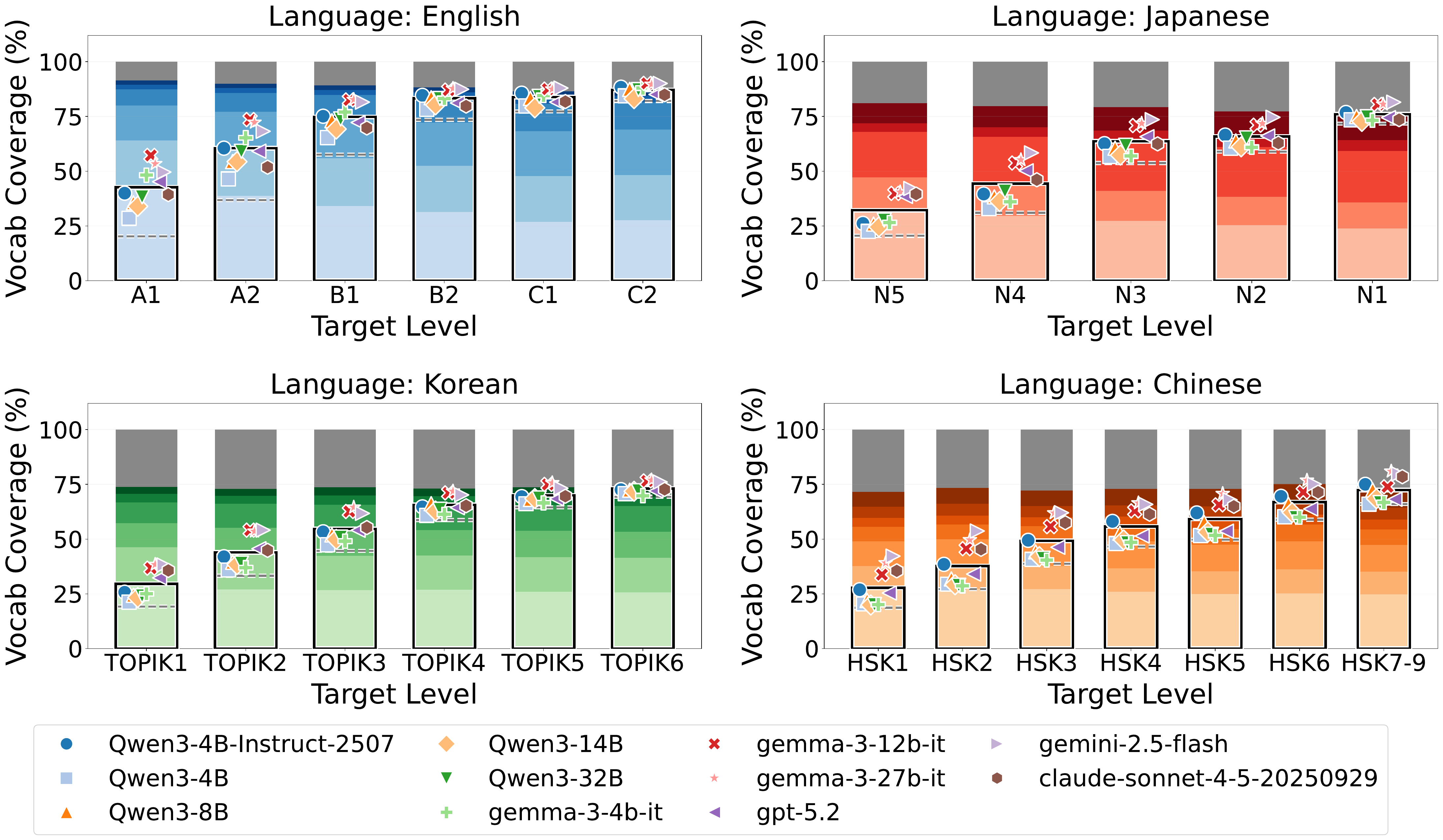}
    \caption{Vocabulary coverage scores (percentage of content words generated at or below the target level) across models and target level variants for each language. The results show that as the target vocabulary level becomes easier, almost all models, including GPT-5.2, Gemini 2.5, and Claude 4.5, fail to provide high vocabulary coverage via the prompting-based approach. The black line box means the average score across models, while the gray dotted line indicates the baseline score of the reference texts. The color depth in each bar plot also represents average vocabulary coverage of each proficiency level from model-generated results for that language (i.e., the lightest color stands for the easiest level), and the gray parts on the top indicate unknown words beyond all levels.}
    \label{fig:zero_shot_vocab_coverage}
\end{figure*}

\section{Can Current LLMs Generate Level-Appropriate Texts?}
\label{sec:can_current_llms_generate_level_appropriate_texts}

We first investigate the performance of existing state-of-the-art LLMs in text simplification at a given learner's vocabulary level.

\subsection{Methodology}
\label{sec:methodology_current_llms}

We compare vocabulary coverage scores for model-generated candidate texts from various LLMs including GPT-5.2, Gemini-2.5-Flash, Claude Sonnet 4.5, Gemma-3, and Qwen3.\footnote{In this paper, all models operate in non-reasoning mode.} For example, if the target level is CEFR B1, we measure what percentage of the total content words are at or below the B1 level (elaborated in \S\ref{sec:vocabulary_coverage}). The models are zero-shot prompted with the language, target proficiency level, and an original text in the Wikipedia dataset from \S\ref{sec:non_parallel_training_data_construction}. The prompt is as shown in Table~\ref{tab:text_simplification_test_prompt}.

\begin{table}[tb]
    \centering
    \small
    \begin{tabular}{p{0.9\linewidth}}
        \toprule
        \textbf{Text Simplification Test} \\
        \midrule
        You are a careful rewrite assistant.\\
        Rewrite the <TEXT> in \verb|{language}| so that every word, except proper nouns or proper adjectives, is at or below the \verb|{level}| vocabulary level.\\
        Replace or simplify any other words above \verb|{level}| level with easier alternatives while preserving the original meaning and coherence.\\
        Do not skip, shorten, or omit any part of the text. Keep sentence count and structure.\\
        Output only the fully converted text with no explanations, instructions, or extra words.\\
        \midrule
        <TEXT>\\
        \verb|{original_text}|\\
        \bottomrule
    \end{tabular}
    \caption{Text simplification prompt used in \S\ref{sec:can_current_llms_generate_level_appropriate_texts} and \S\ref{sec:rl_based_improvement_experiments}. We provide LLMs with the original text along with the target language and proficiency level, requesting text simplification that preserves the original meaning.}
    \label{tab:text_simplification_test_prompt}
\end{table}

\subsection{Results}

The experimental models struggle to generate level-appropriate outputs. Figure~\ref{fig:zero_shot_vocab_coverage} demonstrates the results for overall vocabulary coverage scores.

\paragraph{Performance Degradation at Easy Levels.}

For most languages and models, the vocabulary coverage score tends to decrease significantly as the target level becomes easier. This phenomenon implies that even state-of-the-art LLMs lack the ability to generate text that precisely aligns with the learner's vocabulary knowledge while keeping original information as required by the prompt in Table~\ref{tab:text_simplification_test_prompt}. Notably, even GPT-5.2 shows low performance at the English CEFR A1 level, only scoring 45.1\% of vocabulary coverage.

\paragraph{Performance Discrepancies in Languages.}

Non-English languages generally exhibit lower text simplification performance. At the easiest English proficiency level (A1), the models' average score reaches 42.6\%, which is 22.4 points higher than the original text (20.2\%). However, at the easiest Korean proficiency level (TOPIK1), it achieves 29.8\%, only 10.3 points higher than the baseline (19.5\%). The same pattern holds for Japanese (11.8 point gap) and Chinese (9.3 point gap). This tendency is also observed at intermediate proficiency levels, indicating that language models struggle more with adjusting text difficulty for non-English languages.

\section{\framework: Reward Modules}
\label{sec:reward_modules}

To overcome the limitation in \S\ref{sec:can_current_llms_generate_level_appropriate_texts}, we introduce \framework, a reinforcement learning framework based on the GRPO algorithm~\citep{shao2024deepseekmathpushinglimitsmathematical} to improve LLMs' abilities to perform level-appropriate text simplification. \framework\ integrates three reward modules: vocabulary coverage, semantic preservation, and coherence.

\subsection{Vocabulary Coverage}
\label{sec:vocabulary_coverage}

We design a vocabulary coverage score to calculate whether the vocabulary level of a given text is at or below the target level by matching it with the reference vocabulary level data from \S\ref{sec:multilingual_vocabulary_level_data}.

First, a candidate (simplified) input is lemmatized and processed to remove function words, stopwords, and proper nouns.\footnote{Before the English vocabulary matching, the module conducts phrase-level matching in descending order of length to mitigate the word-level breakdown of phrasal expressions.} Then, the vocabulary coverage score is calculated by matching remaining content lemmas in the input.\footnote{Chinese words are often compound expressions from individual morphological characters. Therefore, if matching fails, they are once more decomposed into individual characters.} Namely, we define the vocabulary coverage score as follows:
\begin{equation}
    \mathrm{score_{vocab}}=\frac{|\{w_i \in M(C) \mid \ell(w_i) \le \ell_t\}|}{|M(C)|},
\label{eq:vocab_coverage}
\end{equation}

where $M(C)$ is a multiset of content lemmas in a candidate input $C$, and $\ell(w_i)$ indicates the proficiency level of a lemma $w_i$, given a target level $\ell_t$. Unknown words not in the reference vocabulary level data are considered as beyond the target level.

To assess improvements from the original text, we then define the final reward $r_{vocab}=\mathrm{score_{vocab\_rollout}}-\mathrm{score_{vocab\_original}}$.\footnote{Since the reward is calculated using only content words, the vocabulary coverage score may not strictly adhere to the linguistically targeted vocabulary coverages. Nevertheless, the score remains useful due to the nature of the GRPO performing relative comparisons.}

\subsection{Semantic Preservation}
\label{sec:semantic_preservation}

We introduce $1:N$ sentence-level bidirectional entailment score for assessing semantic preservation. Through this method, a complex reference sentence can be split into up to $N$ simple corresponding sentences while preserving the meaning.

\begin{figure}[tb]
    \centering
    \includegraphics[width=\columnwidth]{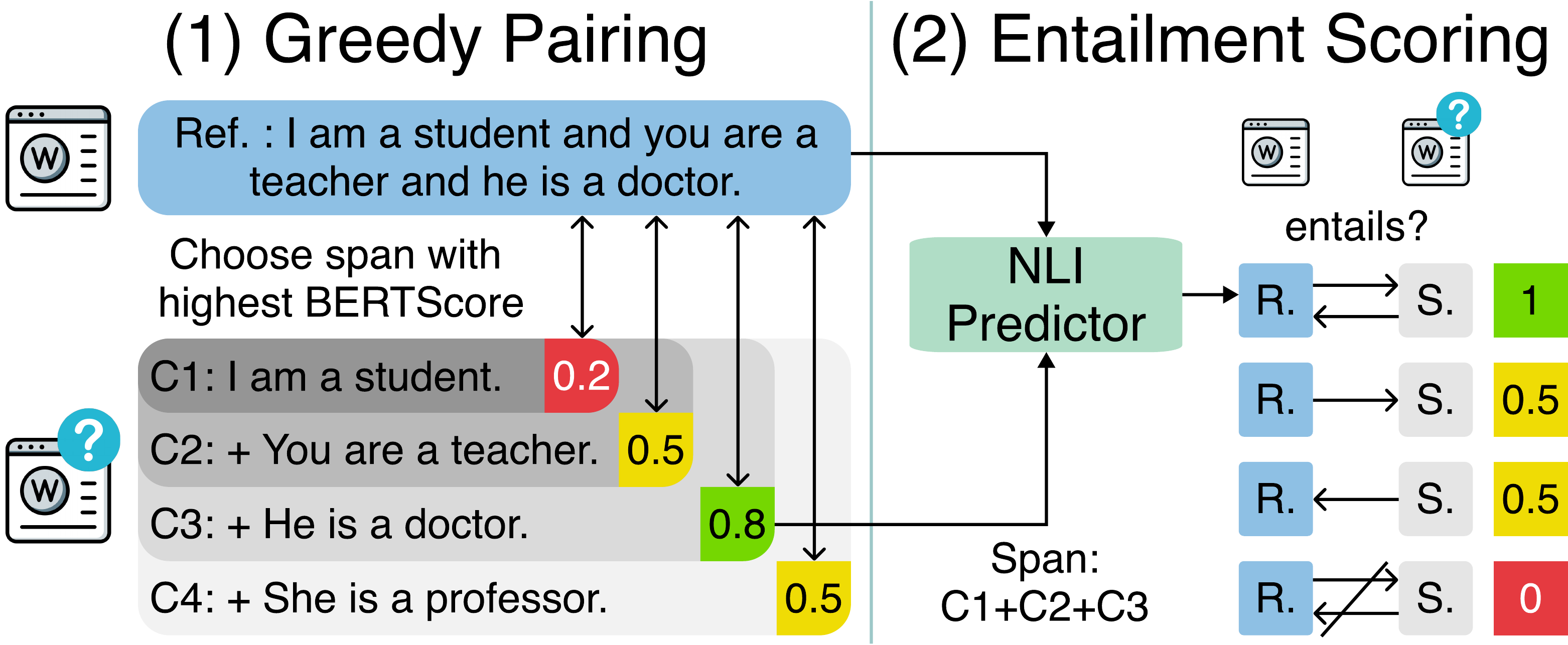}
    \caption{Entailment scoring process for the semantic preservation reward. We introduce a two-phase approach with (1) greedy pairing using BERTScore and (2) entailment scoring via an NLI predictor.}
    \label{fig:entailment}
\end{figure}

To achieve this, we adopt a two-phase approach as shown in Figure~\ref{fig:entailment}. First, for each reference sentence, we greedily combine 1 to $n$ candidate sentences,\footnote{In this paper, the maximum sentence spanning $n=4$.} measuring BERTScore~\citep{zhang2020bertscoreevaluatingtextgeneration} between the reference and candidate sentence spanning pair. We then select the pair with the highest embedding similarity as an aligned result pair. Next, we use a small Natural Language Inference (NLI) model\footnote{We use \texttt{MoritzLaurer/mDeBERTa-v3-base-xnli-\allowbreak multilingual-nli-2mil7} model to determine the sentence entailment.} to bidirectionally examine whether the aligned pairs of a reference sentence $A_i$ and a candidate sentence spanning $B_i$ are in an entailment relationship.

Then, $\mathrm{score_{entailment}}$ is the average of all scores $p_i$ of the aligned pairs, where $p_i$ is defined as:
\begin{equation}
    p_i =
    \begin{cases}
    1.0 & \text{if } A_i \Rightarrow B_i \land B_i \Rightarrow A_i \\
    0.5 & \text{if } A_i \Rightarrow B_i \oplus B_i \Rightarrow A_i \\
    0 & \text{otherwise}
    \end{cases}.
\end{equation}

Finally, the semantic preservation reward $r_{sem\_pres}=\mathrm{score_{entailment}}$.

\subsection{Coherence}

We employ the LLM-as-a-judge~\citep{li-etal-2025-generation} method to measure coherence score by comparatively prompting an evaluator model to assess the naturalness of the generated texts. The evaluator model is assumed to have more parameters than a policy model.

The evaluator model is given a pair of reference and candidate text, with a prompt that requests quality evaluation of the candidate text on a scale from 0 to 100, compared to the reference text. In the prompt, the module requires the evaluator to assign text quality to a predefined score range, heavily penalizing repetitive template phrasing and unnatural sentence patterns that may occur in case of reward hacking. After the evaluation, the coherence score is normalized to a float number between 0 and 1. For the detailed prompt, refer to the Appendix~\ref{sec:prompts}.

During the training, the coherence score is passed through a quadratic transformation to impose a greater penalty on low-quality rollouts. In addition, to prevent the policy model from generating outputs copying the reference text, the coherence module slightly penalizes the score using the Jaccard similarity $J(\cdot)$. The coherence reward $r_{coherence}$ therefore defined as:
\begin{equation}
\begin{aligned}
r_{\mathrm{coherence}}
=
&\max\!\left(1-
\left(
\frac{1-\mathrm{score}_{\mathrm{coherence}}}{1-\alpha}
\right)^2,
0
\right)
\\
&
-
\beta\, J(S_t(A), S_t(B)),
\end{aligned}
\end{equation}

where $\alpha$ is a quality boundary constant,\footnote{$\alpha = 0.6$, which corresponds to the upper boundary of the prompt-defined “machine-generated” quality score.} $\beta$ is a penalty coefficient,\footnote{$\beta = 0.05$, which empirically yields stable training.} and $S_t(A)$ and $S_t(B)$ denote the sets of lemmas in a reference $A$ and a candidate rollout $B$ that exceed the target level $t$, respectively.

\section{\framework: Experiments}
\label{sec:rl_based_improvement_experiments}

\begin{table*}[tb]
    \centering
    \small
    \begin{tabular}{llcccccc}
    \toprule
    \textbf{Lang.} & \textbf{Method} &
    \multicolumn{2}{c}{\makebox[0.28\textwidth][c]{\textbf{Vocabulary Coverage}}} &
    \multicolumn{2}{c}{\makebox[0.14\textwidth][c]{\textbf{Semantic Pres.}}} &
    \multicolumn{2}{c}{\makebox[0.14\textwidth][c]{\textbf{Coherence}}} \\
    \cmidrule(lr){3-4}\cmidrule(lr){5-6}\cmidrule(lr){7-8}
    & & Total (Std.) & Easy (Std.) & Total & Easy & Total & Easy \\
    \midrule
    \multirow{6}{*}{EN}
    & Reference & 58.3 (24.2) & 28.9 (12.6) & -- & -- & -- & -- \\
    & Base (Qwen3-4B) & 72.6 (19.2) & 50.4 (14.4) & 75.3 & 72.3 & 86.9 & 83.7 \\
    & FUDGE (Qwen3-4B) & 74.2 (18.4) & 53.4 (14.4) & 75.0 & 71.8 & 86.7 & 83.1 \\
    & GPT-5.2 & 71.0 (16.8) & 52.4 (12.2) & 76.1 & 70.4 & 84.6 & 75.8 \\
    & Gemini 2.5 & 77.7 (16.3) & 59.1 (13.5) & 72.0 & 69.2 & 82.8 & 78.7 \\
    & \textbf{\framework} & \textbf{81.6} (13.6) & \textbf{66.9} (12.3) & 80.8 & 78.1 & 82.9 & 77.0 \\
    \midrule
    \multirow{6}{*}{JA}
    & Reference & 46.6 (20.3) & 26.4 (9.1) & -- & -- & -- & -- \\
    & Base (Qwen3-4B) & 53.9 (20.4) & 33.7 (10.8) & 65.4 & 64.9 & 88.7 & 88.2 \\
    & FUDGE (Qwen3-4B) & 54.3 (20.3) & 34.2 (10.9) & 64.5 & 65.0 & 88.7 & 88.2 \\
    & GPT-5.2 & 58.8 (15.7) & 45.1 (10.9) & 59.7 & 57.5 & 84.7 & 79.9 \\
    & Gemini 2.5 & 65.8 (16.5) & 51.1 (12.3) & 61.7 & 62.5 & 85.2 & 82.2 \\
    & \textbf{\framework} & \textbf{76.0} (15.3) & \textbf{60.4} (10.7) & 80.6 & 80.1 & 83.1 & 79.6 \\
    \midrule
    \multirow{6}{*}{KO}
    & Reference & 48.8 (19.9) & 26.3 (11.2) & -- & -- & -- & -- \\
    & Base (Qwen3-4B) & 55.2 (19.3) & 33.9 (13.0) & 75.7 & 74.3 & 89.1 & 89.0 \\
    & FUDGE (Qwen3-4B) & 56.0 (19.2) & 34.9 (13.2) & 75.3 & 73.4 & 89.2 & 88.8 \\
    & GPT-5.2 & 56.3 (17.4) & 39.0 (12.6) & 65.6 & 64.1 & 88.5 & 85.4 \\
    & Gemini 2.5 & 62.4 (16.4) & 46.0 (13.9) & 69.3 & 69.3 & 87.0 & 85.0 \\
    & \textbf{\framework} & \textbf{70.4} (15.7) & \textbf{52.9} (12.6) & 87.1 & 86.7 & 84.4 & 81.3 \\
    \midrule
    \multirow{6}{*}{ZH}
    & Reference & 44.8 (20.3) & 23.9 (9.0) & -- & -- & -- & -- \\
    & Base (Qwen3-4B) & 55.2 (19.4) & 33.3 (10.4) & 62.0 & 61.2 & 90.2 & 89.7 \\
    & FUDGE (Qwen3-4B) & 55.4 (19.4) & 33.5 (10.5) & 61.7 & 61.7 & 90.2 & 89.8 \\
    & GPT-5.2 & 50.3 (19.8) & 31.0 (11.0) & 57.7 & 57.6 & 88.4 & 86.0 \\
    & Gemini 2.5 & 64.4 (15.4) & 48.6 (11.9) & 66.7 & 68.1 & 85.3 & 81.2 \\
    & \textbf{\framework} & \textbf{80.2} (12.2) & \textbf{66.1} (9.7) & 76.6 & 77.1 & 83.7 & 81.0 \\
    \bottomrule
    \end{tabular}
    \caption{Evaluation results across four languages, reporting averaged vocabulary coverage, semantic preservation, and coherence scores for overall proficiency levels (Total) and easier-level subsets (Easy: average over the two lowest levels for each language). \framework\ outperforms baselines in vocabulary coverage while preserving meaning and fluency, only using a 4B policy model. In particular, \framework\ improves text simplification performance at easier levels, while existing baselines (including GPT-5.2 and Gemini 2.5) fail to achieve high vocabulary coverage.}
    \label{tab:main_result}
\end{table*}
\begin{table*}[tb]
    \centering
    \small
    \begin{tabular}{llp{0.72\textwidth}}
    \toprule
    \textbf{Lang.} & \textbf{Level} & \textbf{Text} \\
    \midrule
    \multirow{6}{*}{EN} & Reference & Pigeon photography is an \textbf{aerial photography technique invented} in 1907 by the German apothecary Julius Neubronner, who also used pigeons to deliver medications. ... \\
    & CEFR A2 & Pigeon photography is a way of \textbf{taking photos from the air. It was made} in 1907 by a man named Julius Neubronner from Germany. He also used pigeons to carry medicine. ... \\
    & CEFR B2 & Pigeon photography is a way of \textbf{taking photos from the air that was first created} in 1907 by a German man named Julius Neubronner. He also used pigeons to carry medicine. ... \\
    \midrule
    \multirow{8}{*}{JA} & Reference & ... \begin{CJK}{UTF8}{min}ブリオンは「シューマンにとって、限界があり、厳密さを欠く文章表現よりも、 \end{CJK} ... \begin{CJK}{UTF8}{min} 緻密な言葉を提供した」と述べている。\end{CJK}\\
    & & (\textit{Translation: ... Brion states, ``For Schumann, it offered precise language rather than the limited and imprecise nature of written expression ... ''})\\
    & JLPT N4 & ... \begin{CJK}{UTF8}{min}ブリオンさんはこう言っていた。「シューマンにとって、文章というのは限界があって、とても正確でない場合が多い。 \end{CJK} ... \begin{CJK}{UTF8}{min} よく考えられた言葉をくれる。」\end{CJK}\\
    & JLPT N2 & ... \begin{CJK}{UTF8}{min}ブリオンさんはこう言っている。「シューマンにとって、文章というのは限界があって、とても厳密でないものだった。 \end{CJK} ... \begin{CJK}{UTF8}{min} 正確な言葉を提供してくれた。」\end{CJK}\\
    \midrule
    \multirow{8}{*}{KO} & Reference & ... \begin{CJK}{UTF8}{mj}당시 자유주의자들은 독일인들의 열광적이고 광범위한 지지를 얻으며 세를 넓혀나가고 있었다. 자유주의자들은 독일의 통일을\end{CJK} ... \\
    & & (\textit{Translation: ... At the time, the liberals were expanding their influence, gaining enthusiastic and widespread support from the German people. The liberals sought German unification ...})\\
    & TOPIK2 & ... \begin{CJK}{UTF8}{mj}그때 자유주의 사람들은 독일 사람들의 마음을 많이 얻고, 그 마음을 바탕으로 더 큰 힘을 얻어 갔다. 자유주의 사람들은 독일이 하나가 되는 것을\end{CJK} ... \\
    & TOPIK4 & ... \begin{CJK}{UTF8}{mj}그 시기 자유주의 그룹은 독일 사람들의 매우 강한 관심과 넓은 지지를 받으며 점점 더 큰 영향력을 얻어 갔다. 자유주의 그룹은 독일 전체를 하나의 나라로 만드는\end{CJK} ... \\
    \midrule
    \multirow{9}{*}{ZH} & Reference & ... \begin{CJK}{UTF8}{gbsn}在这种背景下，库克在旅程途中也不时把新发现的地方宣告为英国领土，因此，如果说他的航海旅程完全不具“掠夺”性质，也不是准确的说法。\end{CJK} \\
    & & (\textit{Translation: ... Against this backdrop, Cook periodically declared newly discovered territories as British possessions during his voyages. Therefore, it would be inaccurate to claim that his expeditions were entirely devoid of a ``predatory'' nature.})\\
    & HSK2 & ... \begin{CJK}{UTF8}{gbsn}在这种情况下，库克在去的路上，也常常把新发现的地方说成是英国的地盘。所以，说他整个出海过程一点都没有“拿东西”的意思，其实也不对。\end{CJK} \\
    & HSK4 & ... \begin{CJK}{UTF8}{gbsn}在这种情况下，库克在旅行过程中，也常常把新发现的地方说成是英国的地盘。所以，如果有人说库克的旅行一点都没有“想占便宜”的意思，那其实也不太对。\end{CJK} \\
    \bottomrule
    \end{tabular}
\caption{Examples of proficiency-controlled simplification results across languages. Each block shows the reference and two simplified outputs aligned to different target levels. Without pre-labeled parallel corpora, \framework\ enables the trained 4B policy model to generate level-appropriate sentences, preserving information and coherence.}
\label{tab:main_example}
\end{table*}

\subsection{Baselines}

We set baselines to compare relative performances of \framework. Since the use of multilingual proficiency-labeled parallel corpora falls outside our scope, we exclude methods requiring such data.

\paragraph{Reference Text.}

As the most basic score, the original text is used directly as the result text.

\paragraph{Prompting Approach.}

We measure the evaluation scores for a base policy model (Qwen3-4B-Instruct-2507) and state-of-the-art models (GPT-5.2 and Gemini-2.5-Flash) performing zero-shot text simplification using the prompt in Table~\ref{tab:text_simplification_test_prompt}, as conducted in \S\ref{sec:can_current_llms_generate_level_appropriate_texts}.

\paragraph{Constrained Decoding.}

We also measure scores using a constrained generation method, FUDGE~\citep{yang-klein-2021-fudge}, computing scores for top-100 candidate tokens each step. We use a FUDGE discriminator that predicts the probability of meeting the constraint at the end of the generation. Since a multilingual text level classifier is not available, we instead apply a rule-based approach: at each step, we compute the vocabulary coverage score of the current candidate token and last five tokens.\footnote{The last 5 tokens are considered since the spaCy pipeline and the simplification policy model use different tokenizers.}

\subsection{Methodology}

In the training stage, we adopt Qwen3-4B-Instruct-2507 as a policy model, and Qwen3-14B as an evaluator model. We also apply LoRA~\citep{hu2021loralowrankadaptationlarge} for parameter-efficient training. It should be noted that a single policy model is trained for all languages and proficiency levels.

We use the Wikipedia dataset (\S\ref{sec:non_parallel_training_data_construction}) as a training seed data. For each prompt, eight candidate responses are sampled.\footnote{Due to the computational resource constraints, prompts and outputs are limited to a maximum of 512 tokens.} Then, the GRPO objective is optimized using the three reward modules (vocabulary coverage, semantic preservation, and coherence). The final reward is computed as a weighted sum, enabling the level-appropriate simplification policy training that jointly satisfies the multifactorial goal of the experiment.

In the evaluation stage, we employ gemma-3-27b-it and Qwen3-32B as evaluator models and report the average coherence score of both models to avoid self-evaluation.

\subsection{Results}

Table~\ref{tab:main_result} shows overall results, and Table~\ref{tab:main_example} demonstrates simplified text examples in each language. Since easier proficiency levels lead to particularly low scores as observed in \S\ref{sec:can_current_llms_generate_level_appropriate_texts}, we additionally focus on average scores at ``easier'' proficiency levels, defined as the lowest two levels for each language.\footnote{English: A1 and A2, Japanese: N5 and N4, Korean: TOPIK1 and TOPIK2, and Chinese: HSK1 and HSK2.}

\paragraph{Vocabulary Coverage Improvements.}

\framework\ provides substantially higher vocabulary coverage than the baselines, including at easier proficiency levels. For instance, the vocabulary coverage score of the 4B policy model with \framework\ for English simplification achieves 81.6\% across all levels and 66.9\% at easier levels, surpassing Gemini-2.5-Flash (the highest-performing LLM in this work) by 3.9 points and 7.8 points, respectively. Simultaneously, the semantic preservation score improves by 8.8 points to 80.8, while the coherence score increases by 0.1 points to 82.9. Since lower vocabulary coverage inevitably leads to higher coherence in outputs that closely resemble the reference text, achieving high vocabulary coverage while minimizing the coherence reduction is meaningful.

\paragraph{Effectiveness in Non-English Languages.}

\framework\ also significantly improves vocabulary coverage performance for Japanese, Korean, and Chinese. The average vocabulary coverage improvement over the reference texts for non-English languages is 28.8 points, compared to an 8.5-point increase for the FUDGE baseline and 17.5 points for Gemini-2.5-Flash.

\subsection{Data and Model Generalizability}
\label{sec:data_and_model_generalizability}

\begin{figure}[tb]
    \centering
    \includegraphics[width=\columnwidth]{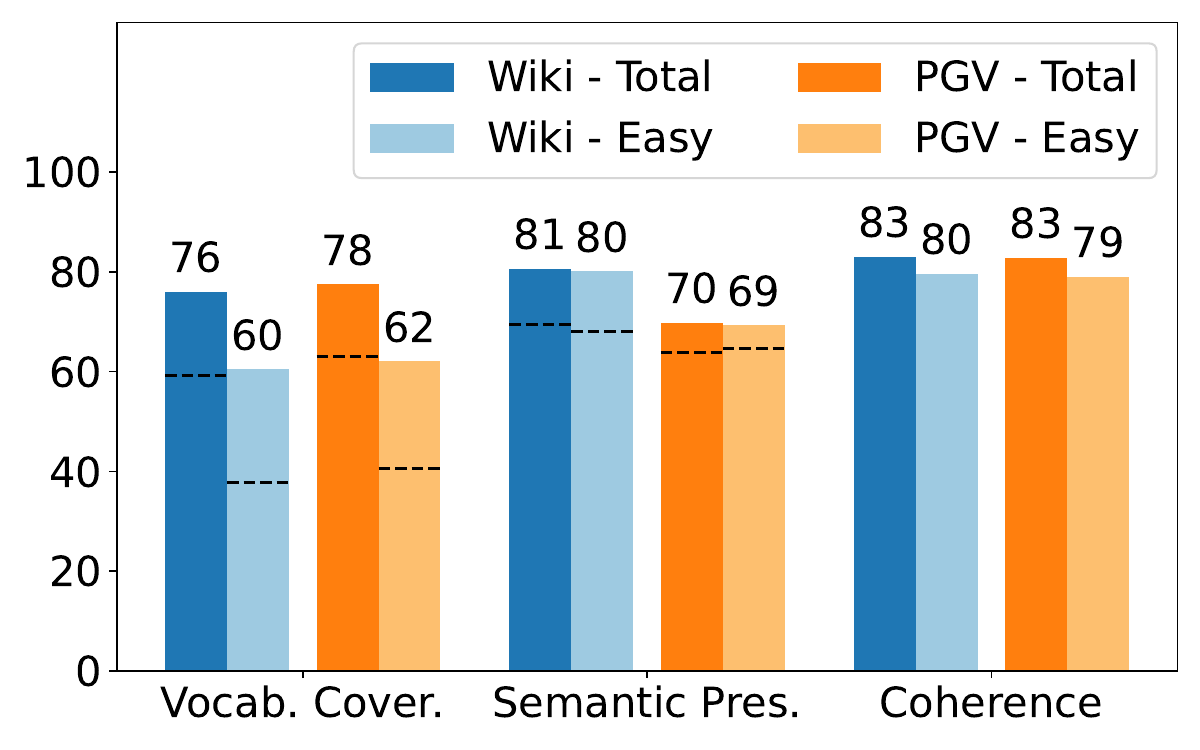}
    \caption{Average evaluation scores across languages on out-of-domain data (PGV), compared to the Wikipedia test data. We show that \framework\ endows the model with general simplification abilities from complex seed texts of an arbitrary domain. The black dashed line represents the performance of the 4B base model.}
    \label{fig:pgv_plot}
\end{figure}

\paragraph{Out-of-Domain Test Set.}
We additionally evaluate our trained policy model on an out-of-domain news dataset from Parallel Global Voices (PGV; elaborated in \S\ref{sec:additional_test_set}) to demonstrate the generalizability of \framework. Figure~\ref{fig:pgv_plot} shows that our framework trained on the Wikipedia dataset effectively operates on PGV test data, implying that sufficiently complex datasets from arbitrary domains can be used as training seed data.

\begin{table}[!htb]
    \centering
    \small
    \begin{tabular}{llccc}
    \toprule
    \textbf{La.} & \textbf{Method} & \textbf{Voc.} & \textbf{Sem.} & \textbf{Coh.}\\
    \midrule
    \multirow{4}{*}{EN}
    & Base (Qw3 4B) & 72.6 & 75.3 & 86.9 \\
    & Base (Ge3 4B) & 74.5 & 69.0 & 84.7 \\
    & \textbf{\framework} (Qw3 4B) & \textbf{81.6} & 80.8 & 82.9 \\
    & \textbf{\framework} (Ge3 4B) & \textbf{81.5} & 79.9 & 80.8 \\
    \midrule
    \multirow{4}{*}{JA}
    & Base (Qw3 4B) & 53.9 & 65.4 & 88.7 \\
    & Base (Ge3 4B) & 50.2 & 61.6 & 90.0 \\
    & \textbf{\framework} (Qw3 4B) & \textbf{76.0} & 80.6 & 83.1 \\
    & \textbf{\framework} (Ge3 4B) & \textbf{72.5} & 76.3 & 79.5 \\
    \midrule
    \multirow{4}{*}{KO}
    & Base (Qw3 4B) & 55.2 & 62.0 & 90.2 \\
    & Base (Ge3 4B) & 51.9 & 69.7 & 90.1 \\
    & \textbf{\framework} (Qw3 4B) & \textbf{70.4} & 87.1 & 84.4 \\
    & \textbf{\framework} (Ge3 4B) & \textbf{72.5} & 84.5 & 82.0 \\
    \midrule
    \multirow{4}{*}{ZH}
    & Base (Qw3 4B) & 55.2 & 62.0 & 90.2 \\
    & Base (Ge3 4B) & 46.3 & 58.8 & 92.1 \\
    & \textbf{\framework} (Qw3 4B) & \textbf{80.2} & 76.6 & 83.7 \\
    & \textbf{\framework} (Ge3 4B) & \textbf{75.1} & 73.6 & 80.3 \\
    \bottomrule
    \end{tabular}
    \caption{Performances of \framework\ on Gemma 3 4B, compared to the baselines with Qwen3 4B. We show that \framework\ can be generalized to other models.}
    \label{tab:generalization_result}
\end{table}

\paragraph{Model Generalization.}
We also demonstrate that \framework\ performs effectively on gemma-3-4b-it. Table~\ref{tab:generalization_result} demonstrates that our framework can be easily integrated into other model families, maintaining the same level of high performance.

\section{Conclusion}

We present \framework, an adaptive, unified, and multilingual text simplification framework tailored to second language learners' proficiency levels, without relying on level-annotated parallel corpora. Our experiments show that even state-of-the-art LLMs have limited abilities to generate level-appropriate outputs, particularly at easy levels and in non-English languages. In contrast, \framework\ substantially improves lexical level controllability while preserving original meaning and fluency. We expect that \framework\ can serve as a foundation for future research toward agentic, adaptive language tutoring systems.

\section*{Limitations}

\paragraph{Scoring Criteria.}

Our vocabulary coverage module requires word level data for each language, where unknown words are treated as words exceeding all proficiency levels. Since it is impossible for researchers to collect all vocabulary used within a language, the proposed scores should not be regarded as absolute values. Nevertheless, the standardized language test criteria require learning a limited vocabulary list for each level, so our methodology can be considered meaningful for measuring vocabulary proficiency levels targeted at language learners.

\paragraph{Granularity of Complexity.}
Additionally, our current approach mainly considers lexical complexity. However, other variables such as syntactic complexity may hold importance in text simplification as well. In this research, we limit the scope of our study to lexical complexity, instead of measuring the hard-to-define syntactic complexity of sentences belonging to different language families.

\paragraph{Model Size Limits.}

Due to the limitation of training resources, we conduct experiments with relatively small 4B-size policy models and a 14B-size evaluator model. This may lead to the generation of slightly unnatural sentences. Moreover, the relatively small NLI predictor we used for the semantic preservation reward may not be able to perfectly determine multilingual entailment relationships. However, the demonstrated generalizability across model families implies that our framework can be readily scaled up to larger models that exhibit more natural multilingual performance.

\paragraph{Needs for Assessment in Real-World Educational Environments.}

Although our framework empirically generates outputs preserving original meaning and naturalness, the evaluations for semantic preservation and coherence are largely substantiated via automated scoring from language models. Therefore, determining the actual usefulness and effects of our framework for language learners in real-world educational settings may require long-term human evaluation experiments.

\section*{Ethical Considerations}

\paragraph{Potential Pedagogical Misuse.}

Although level-appropriate language input can accelerate the second language acquisition process, instructors should not rely solely on automated simplification systems when providing educational materials to language learners. Given the complex nature of language learning, the use of LLM-based readability control frameworks should always be accompanied by appropriate human-in-the-loop review.

\paragraph{Learner Diversity Consideration.}

Language learners exhibit substantial variation in vocabulary knowledge when acquiring foreign languages. While our framework basically assumes learners' lexical proficiency to be aligned with standardized proficiency guidelines, actual vocabulary knowledge may differ depending on learning objectives, educational environments, cultural background, or individual aptitude. Such variation may lead to inappropriate or indiscriminate application of LLM-based frameworks.

\section*{Acknowledgments}

This work was partly supported by Institute of Information \& communications Technology Planning \& Evaluation (IITP) grant funded by the Korean Government (MSIT) (No.~RS-2021-II211343, Artificial Intelligence Graduate School Program (Seoul National University)), the National Research Foundation of Korea(NRF) grant funded by the Korea government(MSIT)(No.~RS-2024-00354218, No.~RS-2024-00353125), and the Technology Innovation Program(RS-2025-25456760, Development of a humanoid robot specialized in chemical processes based on AI foundation model) funded By the Ministry of Trade, Industry and Resources(MOTIR, Korea). We express special thanks to KAIT GPU project. The ICT at Seoul National University provides research facilities for this study.


\bibliography{custom}

\appendix

\section{Experimental Settings}

\subsection{Training Hardware and Software}

\begin{itemize}
    \item Hardware: up to four NVIDIA RTX A6000 GPUs. For proprietary models, we use API calls.
    \item Software: TRL~\citep{vonwerra2022trl} library for the reinforcement learning experiments. Evaluation (including evaluator LLM inferencing during the training process) is conducted using vLLM~\cite{kwon2023efficient} library.
    \item Reproducibility: When needed, the random seed is fixed to 42, and the temperature is set to 0.
\end{itemize}

\subsection{Full List of Language Models Used}

\begin{itemize}
    \item Qwen3-4B-Instruct-2507~\citep{qwen3technicalreport}
    \item Qwen3-4B~\citep{qwen3technicalreport}
    \item Qwen3-8B~\citep{qwen3technicalreport}
    \item Qwen3-14B~\citep{qwen3technicalreport}
    \item Qwen3-32B~\citep{qwen3technicalreport}
    \item gemma-3-4b-it~\citep{gemmateam2025gemma3technicalreport}
    \item gemma-3-12b-it~\citep{gemmateam2025gemma3technicalreport}
    \item gemma-3-27b-it~\citep{gemmateam2025gemma3technicalreport}
    \item gpt-5.2~\citep{openai_gpt52_2025}
    \item gemini-2.5-flash~\citep{comanici2025gemini25pushingfrontier}
    \item claude-sonnet-4-5-20250929~\citep{anthropic_claude_sonnet_4_5_2025}
\end{itemize}

\subsection{Detailed Hyperparameters for \framework\ Experiment}
\label{sec:detailed_hyperparameters}

\begin{itemize}
    \item LoRA configuration.
    \begin{itemize}
        \item Rank: 16
        \item LoRA alpha: 32
        \item Target modules: All attention and MLP layers.
    \end{itemize}
    \item Training configuration.
    \begin{itemize}
        \item Learning rate: 3e-5
        \item Optimizer: adamw\_8bit
        \item Scheduler: linear
        \item Number of sample generations: 8
        \item KL penalty beta: 0.002
        \item Batch size per device: 4
        \item Gradient accumulation steps: 4
        \item GPU hardware: three NVIDIA RTX A6000 GPUs for GRPO training and one NVIDIA RTX A6000 GPU for the evaluator LLM.
    \end{itemize}
    \item Reward module weights.
    \begin{itemize}
        \item Vocabulary coverage: 2.0
        \item Semantic preservation: 1.0
        \item Coherence: 1.0
    \end{itemize}
\end{itemize}

The weight settings are empirically chosen. We observe that assigning a relatively high weight to vocabulary coverage yields stable training.

Due to the training resource constraint, models are trained for 0.25 epochs over the training dataset. Nevertheless, the reward curves have nearly plateaued.

\subsection{Baseline Settings}
\label{sec:baseline_settings}

\begin{itemize}
\item FUDGE: top-100 tokens are considered each generation step. At each time step, vocab coverage score of the candidate token and 5 previous tokens are calculated. The implementation uses the custom logit processor feature of the vLLM library.
\end{itemize}

\section{Dataset Details}
\label{sec:details_on_the_datasets}

\subsection{Vocabulary Level Data}

Table~\ref{tab:vocab_level_list} shows the summary of proficiency level data across English, Japanese, Korean, and Chinese.

When processing Japanese vocabulary data, the Kanji, Hiragana, and Katakana representations of a single word are treated as distinct lemmas sharing the same proficiency level. This encourages policy models to replace relatively difficult Kanji (Sino-Japanese) terms with easier Hiragana-based expressions.

\begin{table*}[htb]
\centering
\small
\begin{tabular}{lllr}
    \toprule
    \textbf{Lang.} & \textbf{Criterion} & \textbf{Level} & \textbf{\# Lemma} \\
    \midrule
    EN & CEFR & A1, A2, B1, B2, C1, C2 & 9413 \\
    JA & JLPT & N5, N4, N3, N2, N1 & 14146 \\
    KO & TOPIK & Level 1, 2, 3, 4, 5, 6 & 9357 \\
    ZH & HSK 3.0 & Level 1, 2, 3, 4, 5, 6, 7-9 & 10870 \\
    \midrule
    \textbf{Total} & & & \textbf{43786} \\
    \bottomrule
\end{tabular}
\caption{Overall summary for our vocabulary level data. We collect proficiency level information of lemmas across four languages (English, Japanese, Korean, and Chinese) based on authoritative examinational standards. Vocabulary coverage module utilizes this data to calculate the proficiency level adequacy of an input.}
\label{tab:vocab_level_list}
\end{table*}

\begin{table*}[htb]
    \centering
    \small
    \begin{tabular}{lp{0.72\textwidth}}
    \toprule
    \textbf{Dataset} & \textbf{Preprocessing} \\
    \midrule
    Wikipedia & \begin{itemize}
    \vspace*{-4ex}
    \item Document Filtering \& Chunking
    \begin{itemize}
        \item Extract `Featured articles' documents
        \item Detect paragraphs by double newline characters
        \item Remove reference paragraphs or paragraphs with $<20$ tokens
        \item Ensure each chunk is $<=512$ tokens 
        \item Maintain paragraph boundaries
    \end{itemize}
    \item Sampling
    \begin{itemize}
        \item Duplicate each chunk by number of target levels
        \item Uniformly sample from each language
    \end{itemize} 
\end{itemize} \\
\midrule
    PGV & \begin{itemize}
    \vspace*{-4ex}
    \item Document Filtering \& Chunking 
    \begin{itemize}
        \item Remove paragraphs with key `crawlinfo' 
        \item Remove paragraphs with key `type' and value `title', `caption', `contributor', `notes', or `tweet-info' 
        \item Concatenate all paragraphs with blank in between 
        \item Remove documents with $<300$ tokens 
        \item Truncate to retain $<=512$ tokens only 
    \end{itemize}
    \item Sampling 
    \begin{itemize}
        \item Duplicate each chunk by number of target levels 
        \item Uniformly sample from each language 
    \end{itemize}
\end{itemize} \\
    \bottomrule

    \end{tabular}

\caption{Checklist for dataset preprocessing.}
\label{tab:checklist_preprocessing}
\end{table*}

\subsection{Wikipedia Dataset}
\label{sec:wikipedia_dataset}

Wikipedia's ``Featured Articles'' are the best documents of Wikipedia manually selected by editors, which can be regarded as a well-structured seed data that covers professional subjects with high-level vocabularies~\citep{xu2011measuring}. As described in Table~\ref{tab:checklist_preprocessing}, we extract the Featured Articles from English (6,793 rticles), Japanese (100 articles), Korean (140 articles), and Chinese (1,023 articles) Wikipedia, resulting in a total of 8,057 documents. As described in \S\ref{sec:non_parallel_training_data_construction}, we preprocess the data to be plain texts and chunk up to 512 tokens while maintaining paragraph information. The paragraph boundaries are determined by two new-line characters. We also remove paragraphs that have under 20 tokens or reference strings, such as `\begin{CJK}{UTF8}{min}参考文献\end{CJK}' or `\begin{CJK}{UTF8}{mj}참고 문헌\end{CJK}'. After preprocessing, we duplicate every chunk by the number of target proficiency levels for each language.\footnote{Each data chunk is replicated across all target proficiency levels, with only the proficiency level and language condition prompts varied; the source text itself remains unchanged.} We then uniformly sampled chunks from each language, making the final training and test dataset. In this way, our resulting dataset is not parallel, nor manually pre-annotated with proficiency levels, and can be substituted with any random dataset that contains sufficiently complex texts. Examples of each chunk are shown in Table~\ref{tab:example_wikipedia_chunk}.

\subsection{Parallel Global Voices Dataset}
\label{sec:parallel_global_voices_dataset}

We use Parallel Global Voices dataset~\citep{PROKOPIDIS16.778} as an out-of-distribution test set. Multilingual corpora of English, Japanese, Korean, and Chinese are used for preprocessing, which is described in Table~\ref{tab:checklist_preprocessing}. The body of each document consists of several xml paragraphs. Since each paragraph of the data usually consists of only a few sentences, we concatenate all paragraphs with a blank in between. We truncate later paragraphs to maintain length within 512 tokens, retaining only 1 chunk per document. Paragraphs containing `crawlinfo' tag (usually in another language), or tag `type' and value `title', `contributor', `notes', `tweet-info', or `caption' are removed. We only retain documents containing at least 300 tokens. Like the Wikipedia dataset, we duplicate every document by target proficiency levels for each language. We then uniformly sample 300 documents for each language, resulting in 1,200 documents. Examples of each chunk are shown in Table~\ref{tab:example_parallel_global_voices_chunk}.

\section{Reward Module Details}

\subsection{Vocabulary Coverage}

We remove stopwords and proper nouns from the total vocabulary count when calculating the vocabulary coverage score. During the scoring process, spaCy performs PoS tagging and removes lemmas that are labeled as ``SYM'', ``PUNCT'', ``SPACE'', ``X'', and ``PROPN''. Additionally, lemmas categorized as ``ADP'', ``AUX'', ``PART'', ``SCONJ'', ``CCONJ'', ``DET'', and ``PRON'' are deleted from the candidate text while processing agglutinative languages, Japanese and Korean. The vocabulary coverage module also matches every lemma with pre-defined stopword lists in each language and excludes them from the text. For English and Chinese, we use pre-defined stopwords in the spaCy library. For Japanese, where no official stopword list exists, we crawl the stopword data from \citet{RanksNL2025JapaneseStopwords}. For Korean, we utilize an educational stopword list from National Institute of Korean Language~\citep{Kim2017IntlKoreanCurriculumPhase4}.

\subsection{Semantic Preservation}

In our $1:N$ sentence-level bidirectional entailment scoring for assessing semantic preservation, we first split all sentences in a reference text and a candidate text. As described in \S\ref{sec:semantic_preservation}, we calculate the entailment score by a two-phase approach: greedy pairing and entailment scoring. Accordingly, the module needs to align the sentence counts to effectively compare the reference and candidate text. When the candidate sentence count is larger than the reference sentence count, the module keeps proceeding with the two phases. If the candidate sentence count is equal to the reference sentence count, there is no need for calculating BERTScore for greedy pairing, so the module proceeds directly to the second phase. Otherwise, it means that the candidate text is so concise that the text could not preserve the original sentence structure, and the module produces no reward for that candidate.

\subsection{Coherence}

We observe that applying a linear penalty to the coherence reward consistently encourages the trained policy model to generate relatively unnatural outputs. To address this issue, we adopt quadratic transformation that attenuates coherence score reduction during training.

\section{Ablation Study}
\label{sec:ablation_study}

We further analyze the effect of each reward module. Effective text simplification requires achieving high vocabulary coverage while minimizing degradation in semantic preservation and coherence. We compare the average evaluation scores of languages and proficiency levels with the following settings:
\begin{itemize}
    \item Vanilla model: Base policy model (Qwen3-4B-Instruct-2507) without training.
    \item Full framework: Vocabulary coverage, semantic preservation, and coherence reward modules.
    \item $1:1$ sentence-level entailment: Vocabulary coverage, semantic preservation (maximum sentence spanning $=1$), and coherence reward modules.
    \item Without semantic preservation: Vocabulary coverage and coherence reward modules only.
    \item Without coherence: Vocabulary coverage and semantic preservation modules only.
\end{itemize}

\begin{table}[htb]
\centering
\small
\begin{tabular}{llccc}
\toprule
\textbf{Setting} & \textbf{Voc.} & \textbf{Sem.} & \textbf{Coh.} \\
\midrule
Vanilla & 59.2 & 69.5 & 88.7 \\
Full framework & 76.0 & 80.6 & 83.1 \\
$1:1$ entailment & 74.0 & 73.3 & 83.7 \\
No Sem. & 84.0 & 52.7 & 83.1 \\
No Coh. & 88.6 & 79.4 & 54.6 \\
\bottomrule
\end{tabular}
\caption{Evaluation results from the ablation study.}
\label{tab:ablation_study}
\end{table}

Ablation experiment results demonstrate the effectiveness of \framework. Compared to the $1:1$ sentence-level entailment setting (maximum sentence spanning $=1$), the full framework (maximum sentence spanning $=4$) achieves a 7.3-point improvement in semantic preservation, indicating that decomposing a complex reference sentence into multiple simpler sentences is beneficial for preserving meaning.

Although the setting without the semantic preservation reward yields higher vocabulary coverage than the full framework, its semantic preservation score drops substantially by 27.9 points. This suggests that the model tends to prioritize lexical simplification without preserving the original meaning of texts. Likewise, removing the coherence reward also produces high vocabulary coverage; however, the coherence score greatly decreases to 54.6, falling into the range defined as ``frequently unnatural'' by the coherence scoring prompt and leading to low-quality outputs.

\section{Further Discussion}

\subsection{Tradeoff between Vocabulary Coverage and Coherence}

Due to the nature of comparative prompting with a reference text and a candidate simplification, coherence scores produced by LLM-based evaluators tend to decrease as the two texts become more dissimilar. This represents a tradeoff with our objective of lexical simplification, and it is inevitable that coherence scores decrease in experiments using relatively small (4B) models, especially in non-English language settings.

Importantly, our data and model generalization experiments (\S\ref{sec:data_and_model_generalizability}) demonstrate that our framework is not dependent on specific models, so researchers may be able to mitigate this issue by employing larger policy and evaluator models.

\subsection{Choice of Evaluation Metrics}

We acknowledge that SARI~\citep{xu-etal-2016-optimizing} and LENS~\citep{maddela-etal-2023-lens} are widely used metrics in the field of text simplification. However, both metrics require parallel reference sentence corpora. Since \framework\ is trained in a fully unsupervised reinforcement learning setting, and no proficiency-specific parallel sentence datasets exist for Japanese, Korean, and Chinese to the best of our knowledge, constructing parallel level-aligned corpora in four languages falls outside the scope of this work. Therefore, we instead mainly focus on actual vocabulary coverage scores grounded in language acquisition theories such as controlled vocabulary approaches.

\subsection{Justifications for Excluding Syntactic Simplification}

We acknowledge that syntactic simplification, in addition to lexical simplification, may also facilitate L2 comprehension. Nevertheless, we can offer both psycholinguistic and methodological justifications regarding our decision to exclude syntactic simplification from the scope of this work.

From a psycholinguistic perspective, our focus on lexical simplification can be supported by the Lexical Bottleneck Hypothesis~\citep{hopp2018bilingual}, which suggests that difficulties in L2 text processing often arise not from insufficient syntactic knowledge per se, but from limited lexical knowledge. In particular, limited lexical knowledge imposes significant cognitive load, preventing learners from effectively allocating resources to syntactic parsing. Conversely, increased lexical knowledge facilitates more efficient syntactic processing and promotes grammar acquisition from language input.

From a methodological perspective, this design choice enables our \framework\ framework to be more effectively applied in multilingual settings. Due to substantial typological variation across languages from different language families, syntactic complexity is difficult to measure in a unified standard across languages. In contrast, lexical complexity can be measured in a more consistent and authoritative manner such as proficiency tests, making our framework more scalable and generalizable for multilingual text simplification.

Nevertheless, it is worth noting that our semantic preservation reward module (\S\ref{sec:semantic_preservation}), which employs a 1:N sentence-level bidirectional entailment scoring mechanism, implicitly promotes syntactic simplification. Under our approach, a single sentence in the original text can be expressed as up to n sentences in the simplified output. This design naturally encourages the policy model to decompose long and syntactically complex sentences into multiple shorter and structurally simpler sentences.

\section{Detailed Results}
\label{sec:detailed_results}

\subsection{Vocabulary Coverage of Prompting-Based Baselines}
The following tables (Table~\ref{tab:detail_vocab_coverage_original}, \ref{tab:detail_vocab_coverage_Qwen3-4B-Instruct-2507}, \ref{tab:detail_vocab_coverage_Qwen3-4B}, \ref{tab:detail_vocab_coverage_Qwen3-8B}, \ref{tab:detail_vocab_coverage_Qwen3-14B}, \ref{tab:detail_vocab_coverage_Qwen3-32B}, \ref{tab:detail_vocab_coverage_gemma-3-4b-it}, \ref{tab:detail_vocab_coverage_gemma-3-12b-it}, \ref{tab:detail_vocab_coverage_gemma-3-27b-it}, \ref{tab:detail_vocab_coverage_gpt-5.2}, \ref{tab:detail_vocab_coverage_gemini-2.5-flash}, and \ref{tab:detail_vocab_coverage_claude-sonnet-4-5-20250929}) illustrate zero-shot vocabulary coverage scores for each experimental model, according to the criteria in Figure~\ref{fig:zero_shot_vocab_coverage}.

\subsection{\framework\ Generations}

Table~\ref{tab:example_generation_english} and \ref{tab:example_generation_korean} are examples of \framework\ generations by proficiency level in English and a non-English language. We use Qwen3-4B-Instruct-2507 as a trained policy model.

\subsection{\framework\ Generations Compared with a Prompting-Based Approach}

We present a case where the prompting-based method fails to achieve high vocabulary coverage and \framework\ successfully provides a sufficient number of level-appropriate words. In particular, base models exhibit low performance at easy proficiency levels. Table~\ref{tab:example_comparing} demonstrates an example case for Japanese JLPT N5 (the easiest proficiency level).

\begin{table}[htb]
\centering
\small
    \begin{tabular}{lp{0.7\columnwidth}}
    \toprule
    Reference & ... \begin{CJK}{UTF8}{min}2億円を超える落札額の産駒は以下の6頭いたが、獲得賞金が落札額を上回ったのはアドマイヤグルーヴのみであった。\end{CJK} ...\\
    & (\textit{Translation: ... Six foals sold for over ¥200 million, but only Admire Groove's career earnings exceeded its winning bid. ...})\\
    \midrule
    Base model & ... \begin{CJK}{UTF8}{min}2億円を超える落札額の馬は以下の6頭あるが、賞金が落札額を上回ったのはアドマイヤグルーヴだけだった。\end{CJK} ...\\
    \midrule
    \framework\ & ... \begin{CJK}{UTF8}{min}その中で、2億円以上で売られた子馬は全部で6頭あったが、実際にもらった賞金の方が売ったお金より多いのは、アドマイヤグルーヴという名前の馬だけだった。\end{CJK} ...\\
    \bottomrule
    \end{tabular}
    \caption{An example comparison between generations from the base 4B model with the prompting method and the trained 4B model with \framework.}
    \label{tab:example_comparing}
\end{table}

In Table~\ref{tab:example_comparing}, the base model directly uses the term ``\begin{CJK}{UTF8}{min}落札額\end{CJK} (\textit{winning bid})'', which is a relatively complex Kango (Sino-Japanese word) for beginners. In contrast, \framework\ efficiently converts the word to much simpler phrases, ``\begin{CJK}{UTF8}{min}売られた\end{CJK} (\textit{were sold})'' and ``\begin{CJK}{UTF8}{min}売ったお金\end{CJK} (\textit{money from the sale})'', while fitting the sentence context.

\subsection{Case Study of Failure}

Despite the powerful performance of \framework, trained models occasionally produce undesirable outputs. We analyze several cases where our framework failed to perform as intended.

\begin{table}[htb]
\centering
\small
    \begin{tabular}{lp{0.7\columnwidth}}
    \toprule
    Reference & Judith Arlene Resnik (April 5, 1949 – January 28, 1986) was an American electrical engineer, software engineer, biomedical engineer, \textbf{pilot} and NASA astronaut who died in the Space Shuttle Challenger disaster.\\
    \midrule
    CEFR A1 & Judith Arlene Resnik (April 5, 1949 – January 28, 1986) was an American person who worked in electricity, software, medicine, \textbf{flew planes}, and also worked for NASA. She died in the Space Shuttle Challenger problem.\\
    \bottomrule
    \end{tabular}
    \caption{An example failure case: grammatical error.}
    \label{tab:failure_1}
\end{table}

\begin{table}[htb]
\centering
\small
    \begin{tabular}{lp{0.7\columnwidth}}
    \toprule
    Reference & \begin{CJK}{UTF8}{mj}어린 프리드리히는 1840년대 이래 독일에서 맹위를 떨치던 자유주의 세력 중심의 혼탁한 정국을 겪었다.\end{CJK}\\
    & (\textit{Translation: Young Friedrich experienced the turbulent political climate dominated by liberal forces that had been rampant in Germany since the 1840s.})\\
    \midrule
    TOPIK1 & \begin{CJK}{UTF8}{mj}어린 프리드리히는 1840년대부터 독일에서 자유주의 세력이 강하게 나타나는 혼란스러운 정치 상황을 겪었다.\end{CJK}\\
    \midrule
    TOPIK6 & \begin{CJK}{UTF8}{mj}어린 프리드리히는 1840년대부터 독일에서 자유주의 세력이 강하게 나타나는 혼란스러운 정치 상황을 겪었다.\end{CJK}\\
    \bottomrule
    \end{tabular}
    \caption{An example failure case: proficiency level misalignment (identical outputs).}
    \label{tab:failure_2}
\end{table}

Table~\ref{tab:failure_1} shows an example case where the trained policy model generates ungrammatical or unnatural outputs. Also, Table~\ref{tab:failure_2} represents a failure case to control proficiency levels, generating simpler but identical sentences for the easiest level (TOPIK1) and the most difficult level (TOPIK6) in Korean. These types of failures are more often observed in non-English generations, where smaller-parameter models can particularly be affected. Therefore, these cases may stem from the small size of the policy model (4B) and evaluator model we employ.

\begin{table*}[htb]
\centering

\begin{subfigure}{\textwidth}
    \centering
    \small
        \begin{tabular}{p{0.9\textwidth}}
            \toprule
                Pigeon photography is an aerial photography technique invented in 1907 by the German apothecary Julius Neubronner, who also used pigeons to deliver medications. A homing pigeon was fitted with an aluminium breast harness to which a lightweight time-delayed miniature camera could be attached. Neubronner's German patent application was initially rejected, but was granted in December 1908 after he produced authenticated photographs taken by his pigeons. He publicized the technique at the 1909 Dresden International Photographic Exhibition, and sold some images as postcards at the Frankfurt International Aviation Exhibition and at the 1910 and 1911 Paris Air Shows.\\\\
                Initially, the military potential of pigeon photography for aerial reconnaissance appeared interesting. Battlefield tests in World War I provided encouraging results, but the ancillary technology of mobile dovecotes for messenger pigeons had the greatest impact. Owing to the rapid development of aviation during the war, military interest in pigeon photography faded and Neubronner abandoned his experiments. The idea was briefly resurrected in the 1930s by a Swiss clockmaker, and reportedly also by the German and French militaries. Although war pigeons were deployed extensively during World War II, it is unclear to what extent, if any, birds were involved in aerial reconnaissance. The United States Central Intelligence Agency (CIA) later developed a battery-powered camera designed for espionage pigeon photography; details of its use remain classified.\\
            \bottomrule
        \end{tabular}
    \caption{An example data chunk from English Wikipedia.}
\end{subfigure}

\vspace{0.5cm} 

\begin{subfigure}{\textwidth}
    \centering
    \small
        \begin{tabular}{p{0.9\textwidth}}
            \toprule
                \begin{CJK}{UTF8}{min}
                シューマンの読書好きは父親譲りで、主として文学と哲学を好んだ。\end{CJK}\\
                \begin{CJK}{UTF8}{min}シューマンは13歳のとき、当時興味を持った批評や詩、哲学的著作からの引用や自作の劇『精神』（未完）からの断章、両親の文章などを「スクランダー」というペンネームで『美しい黄金色の牧場の葉と花』としてまとめている。\end{CJK}\\
                \begin{CJK}{UTF8}{min}また、1825年から1828年の間に書いた自作の文集を「ムルデ河畔のロベルト」というペンネームで『雑録』としてまとめている。このころ、シューマンはゲーテの『ファウスト』をほとんど全部暗記し、友人たちからは「ファウスト」または「メフィスト」などと呼ばれていた。\end{CJK}\\
                \begin{CJK}{UTF8}{min}このほか、シューマンが手がけた文学作品として、コリオランを題材にした合唱付きの悲劇『ランデンドルファー兄弟』や喜劇『レオンハルトとマンテリエ』、ジャン・パウルから影響を受けた『6月の晩と7月の昼間』という小説があるが、いずれも未完である。\end{CJK}\\
                \begin{CJK}{UTF8}{min}シューマンが文学者をめざさず音楽の道を選んだことについて、ブリオンは「シューマンにとって、限界があり、厳密さを欠く文章表現よりも、音楽はずっと豊かで、多様で、陰影があり、緻密な言葉を提供した」と述べている。
                \end{CJK}\\
            \bottomrule
        \end{tabular}
    \caption{An example data chunk from Japanese Wikipedia.}
\end{subfigure}

\vspace{0.5cm} 

\begin{subfigure}{\textwidth}
    \centering
    \small
        \begin{tabular}{p{0.9\textwidth}}
            \toprule
                \begin{CJK}{UTF8}{mj}어린 프리드리히는 1840년대 이래 독일에서 맹위를 떨치던 자유주의 세력 중심의 혼탁한 정국을 겪었다. 당시 자유주의자들은 독일인들의 열광적이고 광범위한 지지를 얻으며 세를 넓혀나가고 있었다. 자유주의자들은 독일의 통일을 희망하였고 입헌군주론자들은 새 헌법을 만들어 모든 인민들의 평등권 보장, 재산 보호, 그리고 기본 인권 보장을 구호로 내세웠다. 즉, 자유주의자들은 인민들의입장을 대변하고 그들의 뜻에 따라 정책을 수립하는 정부를 원하였다. 프리드리히가 17살이 된 1848년, 민족주의자들과 자유주의자들은 독일 전 지역과 서유럽에 걸쳐 대규모의 시위를 주도하였다. 자유주의와 민족주의 세력은 집회와 결사의 자유, 언론의 자유 등의 자유권의 보장과 독일 의회의 수립, 그리고 헌법의 제정을 요구했다. 비록 독일에서의 혁명은 뚜렷한 족적을 남기지는 않았지만, 프리드리히가 어릴 때 목도한 이 자유주의는 훗날 그의 일생에 걸쳐 큰 영향력을 발휘하게 된다.\end{CJK}\\
            \bottomrule
        \end{tabular}
    \caption{An example data chunk from Korean Wikipedia.}
\end{subfigure}

\vspace{0.5cm} 

\begin{subfigure}{\textwidth}
    \centering
    \small
        \begin{tabular}{p{0.9\textwidth}}
            \toprule
                \begin{CJK}{UTF8}{bsmi}在18世紀中期，西方文明對其他文明的影響是西方思想家的辯論焦點之一，不少學者認為西方文明優化了其他文明，但也有學者認為西方文明的入侵腐化了其他文明。庫克的三次航海探索正值這個辯論的高潮，因此他的航海發現或多或少讓西方思想家對地球另一邊鮮為西方所知的文化有稍進一步的了解。不過，庫克對這個命題並不特別關心，從他的周記所見，也不見得出他對「高貴野蠻人」（Noble Savage）這種在當時盛行的看法有特別的興趣。在19世紀，波蘭裔英國小說家約瑟夫·康拉德曾經對歷代航海家和探險家的動機作出比較，他認為庫克以前的航海家和探險家主要以「掠奪」（acquisitive）為動機，而庫克則主要以「科學」（scientific）為動機，因此兩者本質上具有顯著的分別。但有其他學者認為，庫克三次航海旅程的費用要由英國政府動用公帑承擔，這意味出海的計劃和目的受到納稅人監察，在這種背景下，庫克在旅程途中也不時把新發現的地方宣告為英國領土，因此，如果說他的航海旅程完全不具「掠奪」性質，也不是準確的說法。\end{CJK}\\
            \bottomrule
        \end{tabular}
    \caption{An example data chunk from Chinese Wikipedia (traditional Chinese character version).}
\end{subfigure}

\caption{Examples of non-parallel ``Featured Article'' data chunks, excerpted from different articles.}
\label{tab:example_wikipedia_chunk}
\end{table*}

\begin{table*}[htb]
\centering

\begin{subfigure}{\textwidth}
    \centering
    \small
        \begin{tabular}{p{0.9\textwidth}}
            \toprule
                In the months leading to the election of President John Evans Atta Mills, many Ghanaians, including those abroad, feared that a New Democratic Congress (NDC) win would morph into another reign by the party's founder and former military ruler, President Jerry John Rawlings. The transition was smooth, and the relationship between Presidents Rawlings and Mills, seemed cordial.  JJ Rawlings, an unofficial blog site created to highlight the work and thoughts of the former president posted a positive report on May 18, 2009 about President Rawlings' trip to South Africa on behalf of President Mills. It reported: President Mills thanked President Rawlings for representing Ghana at the inauguration and assured him that there would be follow up measures to take advantage of the opportunities for mutually beneficial relations between Ghana and South Africa. But lately, tensions have been rising between President Mills and other members of the NDC party including President Rawlings. Ato Kwamena Dadzie recently wrote in his post, “A President Under Siege“: President Mills is under siege. No doubt about that. He has become such an easy target that an increasing number of people are baying to take a hit at him. What is most intriguing is that the heaviest hitters are people within his own party. The opposition can take a vacation. Just last week, it was the majority leader in parliament, Alban Bagbin. He said the president, who likes to portray himself as a humble sheep, has surrounded himself with aides and confidantes who behave like foxes, hyenas and lions. These people, according to Mr. Bagbin, tend to intimidate and harass anyone who tries to offer some useful counsel to the president.\\
            \bottomrule
        \end{tabular}
    \caption{An example English data chunk.}
\end{subfigure}

\vspace{0.5cm} 

\begin{subfigure}{\textwidth}
    \centering
    \small
        \begin{tabular}{p{0.9\textwidth}}
            \toprule
                \begin{CJK}{UTF8}{min}
                国際女性デーが世界中で祝われた数日前、女性として初めて宇宙に行ったワレンチナ・テレシコワが、75歳の誕生日を迎えた。 宇宙開発競争の発端は、アメリカとソ連がロケット技術開発でしのぎを削った第二次世界大戦後に遡るが、1957年にソ連が人工衛星スプートニクを打上げたことで本格的な競争が始まった。1961年には、ボストーク（ロシア語で東の意味）1号に搭乗したユーリイ・ガガーリンが人類として初めて宇宙に行った。数年後、ボストーク6号が打上られ、テレシコワが宇宙飛行した初の女性となった。 Engineering Pathway Blogでは、去年、テレシコワの宇宙飛行を記念し、彼女の飛行時間の記録に注目した。大気圏に再突入した時点で、テレシコワの飛行時間は、それまでのアメリカ人宇宙飛行士全員の飛行時間の合計を上回っていたのだ。 歴史上の今日：1963年6月16日、ソ連のボストーク6号に搭乗したワレンチナ・テレシコワが初めて宇宙に行った女性となった。3日間の宇宙飛行を終えたテレシコワの飛行時間は、当時のアメリカ人宇宙飛行士全員の飛行時間の合計を上回った。[…]
                \end{CJK}\\
            \bottomrule
        \end{tabular}
    \caption{An example Japanese data chunk.}
\end{subfigure}

\vspace{0.5cm} 

\begin{subfigure}{\textwidth}
    \centering
    \small
        \begin{tabular}{p{0.9\textwidth}}
            \toprule
                \begin{CJK}{UTF8}{mj}해변가의 파라솔 그늘 아래 여유롭게 누워있거나, 관광객이 되어 손에 가이드북을 들고 생소한 도시의 거리를 돌아다니거나, 아니면 집에서 휴식을 취하며 자유시간을 즐기기. 이게 휴일을 보내는 일반적인 방법이 아닐까 싶다. 그러나 많은 일본인들은 직장에서 최대한 휴가를 내지 않으려 이러한 휴일을 매년 날려 버리곤 한다. 익스페디아 (Expedia) 일본지부가 선진국 11개국을 대상으로 한 ‘휴가 부족 (Vacation Deprivation) 상황에 관한 설문조사'에 따르면 일본 노동자들은 1년에 평균 7.9일의 휴가를 보낸다. 일본에서는 평균적으로 15일의 유급 휴가가 주어지는데 이는 미국에 이어 두 번째로 낮은 수다. 사용하지 않은 휴가 일수로는 일본이 가장 높은 것으로 나타났다. 이는 일본인들이 경제적 위기에 중압감과 불안감을 느끼고 있고, 또한 휴가를 사용함으로써 “동료에게 업무 부담을 늘리는 폐'를 끼치지 않고자 하는 직장 내 분위기 때문인 것으로 보여진다.\end{CJK}\\
            \bottomrule
        \end{tabular}
    \caption{An example Korean data chunk.}
\end{subfigure}

\vspace{0.5cm} 

\begin{subfigure}{\textwidth}
    \centering
    \small
        \begin{tabular}{p{0.9\textwidth}}
            \toprule
                \begin{CJK}{UTF8}{gbsn}今日世界看似平坦，从亚洲、非洲、欧洲到美洲，世界人民都感受到全球经济衰退的负面后果，本文试图记录在庞大金融危机之下，世界各地一般民众所见所受的社会冲击。 这场风暴最显著的迹象为美国华尔街重挫与房地产市场崩盘，虽然我们不该小看这些经济灾害，但更要关注全球民众每日目睹与身陷的灾情。 例如经济警讯已迫使许多韩国民众更改或放弃旅游计划，汶莱向来是韩国人的观光胜地，如今旅游业亦受波及；由于旅客人数下滑，埃及观光业人员已抱怨薪水未如期入帐。 美国消费力萎缩压低孟加拉成长出口需求，影响当地雇员众多的成衣出口业，美国与英国为孟加拉产品主要出口国。 俄罗斯金融危机反映在政府削减医疗支出；日本衰退情况包括缩短百货公司营业时间、车辆销量降低，以及临时屋宅、公园及网咖里出现愈来愈多失业游民。 经济急冻在乌克兰非常具体，去年12月由于政府未准时缴纳水费，使首都基辅多数地区一星期无热水可用，Evie of Kiva Stories from the Field博客描述基辅居民受冻的困境： 在零下低温与大陆型寒冬夹击下，一星期没有热水着实痛苦，人们甚至无法洗碗盘，因为水全都在管线里冻结，尽管热水恢复供应三天后，暖气装置功能尚未正常，住家仍然寒冷，民众也受到风寒感冒。 香港向来以全球金融中心见长，很意外得知汇丰控股股价重挫至1995年来的最低点，更让一位电视评论员在提及此事时，因不可置信而落泪。\end{CJK}\\
            \bottomrule
        \end{tabular}
    \caption{An example Chinese data chunk.}
\end{subfigure}

\caption{Examples of Parallel Global Voices data chunks, excerpted from different articles.}
\label{tab:example_parallel_global_voices_chunk}
\end{table*}

\begin{table}[htb]
\centering
\small
\begin{tabular}{llc}
\toprule
\textbf{Lang.} & \textbf{Level} & \textbf{Original} \\
& & (Avg. $\pm$ Std.) \\
\midrule
EN & A1 & 20.2 $\pm$ 7.5 \\
 & A2 & 37.5 $\pm$ 10.5 \\
 & B1 & 58.0 $\pm$ 10.4 \\
 & B2 & 73.5 $\pm$ 9.0 \\
 & C1 & 77.2 $\pm$ 8.5 \\
 & C2 & 81.9 $\pm$ 7.6 \\
\midrule
JA & N5 & 20.6 $\pm$ 6.3 \\
 & N4 & 31.3 $\pm$ 8.2 \\
 & N3 & 53.5 $\pm$ 10.2 \\
 & N2 & 58.8 $\pm$ 9.5 \\
 & N1 & 71.6 $\pm$ 10.2 \\
\midrule
KO & TOPIK1 & 19.5 $\pm$ 8.6 \\
 & TOPIK2 & 33.4 $\pm$ 9.1 \\
 & TOPIK3 & 44.4 $\pm$ 9.9 \\
 & TOPIK4 & 58.0 $\pm$ 10.4 \\
 & TOPIK5 & 63.9 $\pm$ 10.3 \\
 & TOPIK6 & 68.9 $\pm$ 10.6 \\
\midrule
ZH & HSK1 & 19.1 $\pm$ 6.0 \\
 & HSK2 & 28.2 $\pm$ 9.1 \\
 & HSK3 & 40.0 $\pm$ 11.7 \\
 & HSK4 & 47.4 $\pm$ 12.2 \\
 & HSK5 & 51.2 $\pm$ 14.8 \\
 & HSK6 & 59.3 $\pm$ 15.5 \\
 & HSK7-9 & 67.0 $\pm$ 19.3 \\
\bottomrule
\end{tabular}
\caption{Vocabulary coverage score baseline statistics for Reference Texts in Wikipedia Featured Article Dataset.}
\label{tab:detail_vocab_coverage_original}
\end{table}

\begin{table}[htb]
\centering
\small
\begin{tabular}{llc}
\toprule
\textbf{Lang.} & \textbf{Level} & \textbf{Simplified} \\
& & (Avg. $\pm$ Std.) \\
\midrule
EN & A1 & 39.8 $\pm$ 9.7 \\
 & A2 & 60.9 $\pm$ 9.8 \\
 & B1 & 75.0 $\pm$ 10.0 \\
 & B2 & 84.8 $\pm$ 8.2 \\
 & C1 & 85.5 $\pm$ 7.2 \\
 & C2 & 88.3 $\pm$ 6.6 \\
\midrule
JA & N5 & 26.4 $\pm$ 7.6 \\
 & N4 & 39.8 $\pm$ 9.3 \\
 & N3 & 62.4 $\pm$ 10.2 \\
 & N2 & 66.2 $\pm$ 9.9 \\
 & N1 & 76.9 $\pm$ 9.9 \\
\midrule
KO & TOPIK1 & 25.9 $\pm$ 9.8 \\
 & TOPIK2 & 42.0 $\pm$ 10.6 \\
 & TOPIK3 & 53.0 $\pm$ 10.8 \\
 & TOPIK4 & 64.5 $\pm$ 10.5 \\
 & TOPIK5 & 69.0 $\pm$ 10.5 \\
 & TOPIK6 & 72.4 $\pm$ 10.9 \\
\midrule
ZH & HSK1 & 27.4 $\pm$ 7.6 \\
 & HSK2 & 38.9 $\pm$ 9.6 \\
 & HSK3 & 50.8 $\pm$ 11.8 \\
 & HSK4 & 58.7 $\pm$ 10.3 \\
 & HSK5 & 62.7 $\pm$ 12.4 \\
 & HSK6 & 70.2 $\pm$ 12.2 \\
 & HSK7-9 & 76.3 $\pm$ 14.1 \\
\bottomrule
\end{tabular}
\caption{Vocabulary coverage score statistics for Qwen3-4B-Instruct-2507.}
\label{tab:detail_vocab_coverage_Qwen3-4B-Instruct-2507}
\end{table}

\begin{table}[htb]
\centering
\small
\begin{tabular}{llc}
\toprule
\textbf{Lang.} & \textbf{Level} & \textbf{Simplified} \\
& & (Avg. $\pm$ Std.) \\
\midrule
EN & A1 & 29.0 $\pm$ 10.9 \\
 & A2 & 47.4 $\pm$ 12.0 \\
 & B1 & 65.5 $\pm$ 11.6 \\
 & B2 & 78.7 $\pm$ 10.1 \\
 & C1 & 79.8 $\pm$ 8.9 \\
 & C2 & 84.4 $\pm$ 7.8 \\
\midrule
JA & N5 & 22.9 $\pm$ 7.1 \\
 & N4 & 34.0 $\pm$ 9.7 \\
 & N3 & 56.7 $\pm$ 10.7 \\
 & N2 & 61.4 $\pm$ 10.3 \\
 & N1 & 73.6 $\pm$ 11.3 \\
\midrule
KO & TOPIK1 & 21.7 $\pm$ 9.5 \\
 & TOPIK2 & 36.2 $\pm$ 9.5 \\
 & TOPIK3 & 47.4 $\pm$ 10.2 \\
 & TOPIK4 & 60.7 $\pm$ 10.4 \\
 & TOPIK5 & 65.9 $\pm$ 10.4 \\
 & TOPIK6 & 70.5 $\pm$ 10.6 \\
\midrule
ZH & HSK1 & 21.1 $\pm$ 7.2 \\
 & HSK2 & 30.7 $\pm$ 10.1 \\
 & HSK3 & 42.3 $\pm$ 13.1 \\
 & HSK4 & 49.3 $\pm$ 13.2 \\
 & HSK5 & 53.3 $\pm$ 16.0 \\
 & HSK6 & 61.0 $\pm$ 15.9 \\
 & HSK7-9 & 67.4 $\pm$ 20.1 \\
\bottomrule
\end{tabular}
\caption{Vocabulary coverage score statistics for Qwen3-4B.}
\label{tab:detail_vocab_coverage_Qwen3-4B}
\end{table}

\begin{table}[htb]
\centering
\small
\begin{tabular}{llc}
\toprule
\textbf{Lang.} & \textbf{Level} & \textbf{Simplified} \\
& & (Avg. $\pm$ Std.) \\
\midrule
EN & A1 & 35.3 $\pm$ 11.1 \\
 & A2 & 54.9 $\pm$ 11.3 \\
 & B1 & 72.9 $\pm$ 11.3 \\
 & B2 & 83.6 $\pm$ 9.1 \\
 & C1 & 83.2 $\pm$ 8.3 \\
 & C2 & 87.2 $\pm$ 7.4 \\
\midrule
JA & N5 & 26.1 $\pm$ 8.3 \\
 & N4 & 38.6 $\pm$ 10.4 \\
 & N3 & 60.5 $\pm$ 10.8 \\
 & N2 & 63.8 $\pm$ 9.9 \\
 & N1 & 75.1 $\pm$ 10.0 \\
\midrule
KO & TOPIK1 & 24.6 $\pm$ 9.7 \\
 & TOPIK2 & 39.4 $\pm$ 10.0 \\
 & TOPIK3 & 51.9 $\pm$ 10.2 \\
 & TOPIK4 & 63.2 $\pm$ 11.6 \\
 & TOPIK5 & 68.4 $\pm$ 10.9 \\
 & TOPIK6 & 72.2 $\pm$ 10.7 \\
\midrule
ZH & HSK1 & 22.6 $\pm$ 6.8 \\
 & HSK2 & 32.6 $\pm$ 9.5 \\
 & HSK3 & 45.6 $\pm$ 12.0 \\
 & HSK4 & 53.5 $\pm$ 11.3 \\
 & HSK5 & 58.3 $\pm$ 13.2 \\
 & HSK6 & 66.6 $\pm$ 12.8 \\
 & HSK7-9 & 73.6 $\pm$ 15.3 \\
\bottomrule
\end{tabular}
\caption{Vocabulary coverage score statistics for Qwen3-8B.}
\label{tab:detail_vocab_coverage_Qwen3-8B}
\end{table}

\begin{table}[htb]
\centering
\small
\begin{tabular}{llc}
\toprule
\textbf{Lang.} & \textbf{Level} & \textbf{Simplified} \\
& & (Avg. $\pm$ Std.) \\
\midrule
EN & A1 & 33.9 $\pm$ 11.0 \\
 & A2 & 54.6 $\pm$ 11.8 \\
 & B1 & 69.5 $\pm$ 11.2 \\
 & B2 & 80.4 $\pm$ 9.8 \\
 & C1 & 78.9 $\pm$ 8.8 \\
 & C2 & 83.0 $\pm$ 7.7 \\
\midrule
JA & N5 & 24.7 $\pm$ 7.5 \\
 & N4 & 36.5 $\pm$ 9.5 \\
 & N3 & 57.1 $\pm$ 10.8 \\
 & N2 & 61.1 $\pm$ 9.8 \\
 & N1 & 72.7 $\pm$ 10.3 \\
\midrule
KO & TOPIK1 & 23.8 $\pm$ 9.7 \\
 & TOPIK2 & 38.6 $\pm$ 10.5 \\
 & TOPIK3 & 49.7 $\pm$ 11.0 \\
 & TOPIK4 & 62.4 $\pm$ 10.6 \\
 & TOPIK5 & 68.0 $\pm$ 10.5 \\
 & TOPIK6 & 71.6 $\pm$ 10.5 \\
\midrule
ZH & HSK1 & 20.5 $\pm$ 6.6 \\
 & HSK2 & 30.4 $\pm$ 9.7 \\
 & HSK3 & 43.6 $\pm$ 12.4 \\
 & HSK4 & 50.8 $\pm$ 12.7 \\
 & HSK5 & 54.6 $\pm$ 15.3 \\
 & HSK6 & 63.0 $\pm$ 15.7 \\
 & HSK7-9 & 70.0 $\pm$ 18.9 \\
\bottomrule
\end{tabular}
\caption{Vocabulary coverage score statistics for Qwen3-14B.}
\label{tab:detail_vocab_coverage_Qwen3-14B}
\end{table}

\begin{table}[htb]
\centering
\small
\begin{tabular}{llc}
\toprule
\textbf{Lang.} & \textbf{Level} & \textbf{Simplified} \\
& & (Avg. $\pm$ Std.) \\
\midrule
EN & A1 & 38.3 $\pm$ 9.6 \\
 & A2 & 59.6 $\pm$ 9.4 \\
 & B1 & 73.9 $\pm$ 10.2 \\
 & B2 & 83.2 $\pm$ 8.9 \\
 & C1 & 84.1 $\pm$ 8.0 \\
 & C2 & 87.2 $\pm$ 7.0 \\
\midrule
JA & N5 & 28.0 $\pm$ 7.7 \\
 & N4 & 41.4 $\pm$ 9.6 \\
 & N3 & 61.7 $\pm$ 10.3 \\
 & N2 & 65.2 $\pm$ 9.6 \\
 & N1 & 74.9 $\pm$ 9.9 \\
\midrule
KO & TOPIK1 & 24.3 $\pm$ 9.8 \\
 & TOPIK2 & 39.3 $\pm$ 9.4 \\
 & TOPIK3 & 51.1 $\pm$ 10.7 \\
 & TOPIK4 & 63.3 $\pm$ 10.6 \\
 & TOPIK5 & 68.4 $\pm$ 10.1 \\
 & TOPIK6 & 71.6 $\pm$ 11.2 \\
\midrule
ZH & HSK1 & 20.8 $\pm$ 6.8 \\
 & HSK2 & 30.2 $\pm$ 9.9 \\
 & HSK3 & 42.8 $\pm$ 12.6 \\
 & HSK4 & 49.6 $\pm$ 13.1 \\
 & HSK5 & 53.8 $\pm$ 15.3 \\
 & HSK6 & 60.8 $\pm$ 16.2 \\
 & HSK7-9 & 67.8 $\pm$ 19.7 \\
\bottomrule
\end{tabular}
\caption{Vocabulary coverage score statistics for Qwen3-32B.}
\label{tab:detail_vocab_coverage_Qwen3-32B}
\end{table}

\begin{table}[htb]
\centering
\small
\begin{tabular}{llc}
\toprule
\textbf{Lang.} & \textbf{Level} & \textbf{Simplified} \\
& & (Avg. $\pm$ Std.) \\
\midrule
EN & A1 & 48.3 $\pm$ 10.4 \\
 & A2 & 65.6 $\pm$ 9.1 \\
 & B1 & 76.9 $\pm$ 9.1 \\
 & B2 & 83.0 $\pm$ 7.6 \\
 & C1 & 84.4 $\pm$ 6.8 \\
 & C2 & 87.8 $\pm$ 5.7 \\
\midrule
JA & N5 & 26.5 $\pm$ 8.5 \\
 & N4 & 36.6 $\pm$ 9.8 \\
 & N3 & 56.4 $\pm$ 10.4 \\
 & N2 & 60.8 $\pm$ 9.7 \\
 & N1 & 73.2 $\pm$ 10.0 \\
\midrule
KO & TOPIK1 & 25.3 $\pm$ 10.1 \\
 & TOPIK2 & 37.1 $\pm$ 9.7 \\
 & TOPIK3 & 48.7 $\pm$ 10.3 \\
 & TOPIK4 & 60.8 $\pm$ 11.1 \\
 & TOPIK5 & 65.8 $\pm$ 10.7 \\
 & TOPIK6 & 69.6 $\pm$ 10.6 \\
\midrule
ZH & HSK1 & 20.7 $\pm$ 7.0 \\
 & HSK2 & 29.8 $\pm$ 9.8 \\
 & HSK3 & 41.8 $\pm$ 12.6 \\
 & HSK4 & 49.5 $\pm$ 13.1 \\
 & HSK5 & 52.9 $\pm$ 15.0 \\
 & HSK6 & 60.6 $\pm$ 16.1 \\
 & HSK7-9 & 67.9 $\pm$ 19.8 \\
\bottomrule
\end{tabular}
\caption{Vocabulary coverage score statistics for gemma-3-4b-it.}
\label{tab:detail_vocab_coverage_gemma-3-4b-it}
\end{table}

\begin{table}[htb]
\centering
\small
\begin{tabular}{llc}
\toprule
\textbf{Lang.} & \textbf{Level} & \textbf{Simplified} \\
& & (Avg. $\pm$ Std.) \\
\midrule
EN & A1 & 57.0 $\pm$ 10.3 \\
 & A2 & 73.8 $\pm$ 8.7 \\
 & B1 & 82.5 $\pm$ 8.3 \\
 & B2 & 87.1 $\pm$ 7.3 \\
 & C1 & 87.2 $\pm$ 6.7 \\
 & C2 & 90.1 $\pm$ 6.2 \\
\midrule
JA & N5 & 40.2 $\pm$ 9.5 \\
 & N4 & 53.8 $\pm$ 10.1 \\
 & N3 & 70.7 $\pm$ 9.8 \\
 & N2 & 70.8 $\pm$ 9.7 \\
 & N1 & 80.4 $\pm$ 9.5 \\
\midrule
KO & TOPIK1 & 36.7 $\pm$ 11.0 \\
 & TOPIK2 & 53.9 $\pm$ 10.2 \\
 & TOPIK3 & 62.4 $\pm$ 11.0 \\
 & TOPIK4 & 70.8 $\pm$ 10.5 \\
 & TOPIK5 & 74.5 $\pm$ 9.4 \\
 & TOPIK6 & 76.4 $\pm$ 10.2 \\
\midrule
ZH & HSK1 & 34.8 $\pm$ 10.7 \\
 & HSK2 & 47.0 $\pm$ 13.9 \\
 & HSK3 & 57.6 $\pm$ 14.4 \\
 & HSK4 & 64.1 $\pm$ 13.3 \\
 & HSK5 & 67.1 $\pm$ 15.0 \\
 & HSK6 & 73.4 $\pm$ 15.0 \\
 & HSK7-9 & 75.3 $\pm$ 18.1 \\
\bottomrule
\end{tabular}
\caption{Vocabulary coverage score statistics for gemma-3-12b-it.}
\label{tab:detail_vocab_coverage_gemma-3-12b-it}
\end{table}

\begin{table}[htb]
\centering
\small
\begin{tabular}{llc}
\toprule
\textbf{Lang.} & \textbf{Level} & \textbf{Simplified} \\
& & (Avg. $\pm$ Std.) \\
\midrule
EN & A1 & 53.4 $\pm$ 10.6 \\
 & A2 & 72.4 $\pm$ 9.0 \\
 & B1 & 82.5 $\pm$ 8.4 \\
 & B2 & 87.8 $\pm$ 6.7 \\
 & C1 & 87.3 $\pm$ 6.2 \\
 & C2 & 89.7 $\pm$ 5.6 \\
\midrule
JA & N5 & 40.7 $\pm$ 9.6 \\
 & N4 & 55.2 $\pm$ 10.0 \\
 & N3 & 71.8 $\pm$ 9.5 \\
 & N2 & 71.6 $\pm$ 9.6 \\
 & N1 & 80.6 $\pm$ 9.2 \\
\midrule
KO & TOPIK1 & 38.4 $\pm$ 11.6 \\
 & TOPIK2 & 53.3 $\pm$ 10.5 \\
 & TOPIK3 & 64.4 $\pm$ 11.4 \\
 & TOPIK4 & 71.6 $\pm$ 9.9 \\
 & TOPIK5 & 75.2 $\pm$ 9.4 \\
 & TOPIK6 & 76.8 $\pm$ 9.8 \\
\midrule
ZH & HSK1 & 39.7 $\pm$ 10.3 \\
 & HSK2 & 50.0 $\pm$ 12.2 \\
 & HSK3 & 62.5 $\pm$ 11.1 \\
 & HSK4 & 66.8 $\pm$ 10.2 \\
 & HSK5 & 71.1 $\pm$ 10.8 \\
 & HSK6 & 77.0 $\pm$ 9.8 \\
 & HSK7-9 & 81.6 $\pm$ 11.4 \\
\bottomrule
\end{tabular}
\caption{Vocabulary coverage score statistics for gemma-3-27b-it.}
\label{tab:detail_vocab_coverage_gemma-3-27b-it}
\end{table}

\begin{table}[htb]
\centering
\small
\begin{tabular}{llc}
\toprule
\textbf{Lang.} & \textbf{Level} & \textbf{Simplified} \\
& & (Avg. $\pm$ Std.) \\
\midrule
EN & A1 & 45.1 $\pm$ 9.8 \\
 & A2 & 59.7 $\pm$ 9.7 \\
 & B1 & 72.5 $\pm$ 9.6 \\
 & B2 & 81.2 $\pm$ 8.5 \\
 & C1 & 81.5 $\pm$ 7.7 \\
 & C2 & 84.7 $\pm$ 8.7 \\
\midrule
JA & N5 & 38.7 $\pm$ 8.4 \\
 & N4 & 50.5 $\pm$ 9.9 \\
 & N3 & 65.6 $\pm$ 9.8 \\
 & N2 & 66.0 $\pm$ 9.7 \\
 & N1 & 74.4 $\pm$ 10.2 \\
\midrule
KO & TOPIK1 & 32.4 $\pm$ 10.8 \\
 & TOPIK2 & 45.7 $\pm$ 10.6 \\
 & TOPIK3 & 53.7 $\pm$ 11.6 \\
 & TOPIK4 & 64.0 $\pm$ 10.0 \\
 & TOPIK5 & 67.2 $\pm$ 12.0 \\
 & TOPIK6 & 71.1 $\pm$ 11.9 \\
\midrule
ZH & HSK1 & 26.1 $\pm$ 8.2 \\
 & HSK2 & 35.5 $\pm$ 11.3 \\
 & HSK3 & 48.0 $\pm$ 13.2 \\
 & HSK4 & 52.7 $\pm$ 13.4 \\
 & HSK5 & 55.3 $\pm$ 15.6 \\
 & HSK6 & 64.3 $\pm$ 15.5 \\
 & HSK7-9 & 69.2 $\pm$ 19.3 \\
\bottomrule
\end{tabular}
\caption{Vocabulary coverage score statistics for gpt-5.2.}
\label{tab:detail_vocab_coverage_gpt-5.2}
\end{table}

\begin{table}[htb]
\centering
\small
\begin{tabular}{llc}
\toprule
\textbf{Lang.} & \textbf{Level} & \textbf{Simplified} \\
& & (Avg. $\pm$ Std.) \\
\midrule
EN & A1 & 49.6 $\pm$ 10.2 \\
 & A2 & 68.7 $\pm$ 8.9 \\
 & B1 & 81.7 $\pm$ 7.3 \\
 & B2 & 87.6 $\pm$ 7.2 \\
 & C1 & 87.9 $\pm$ 6.5 \\
 & C2 & 90.0 $\pm$ 6.1 \\
\midrule
JA & N5 & 42.3 $\pm$ 9.1 \\
 & N4 & 58.5 $\pm$ 9.6 \\
 & N3 & 73.3 $\pm$ 9.0 \\
 & N2 & 74.2 $\pm$ 9.4 \\
 & N1 & 81.4 $\pm$ 9.2 \\
\midrule
KO & TOPIK1 & 38.3 $\pm$ 11.8 \\
 & TOPIK2 & 53.9 $\pm$ 11.2 \\
 & TOPIK3 & 61.6 $\pm$ 10.6 \\
 & TOPIK4 & 69.5 $\pm$ 10.2 \\
 & TOPIK5 & 72.4 $\pm$ 9.7 \\
 & TOPIK6 & 75.7 $\pm$ 9.5 \\
\midrule
ZH & HSK1 & 42.6 $\pm$ 10.0 \\
 & HSK2 & 54.1 $\pm$ 10.8 \\
 & HSK3 & 62.8 $\pm$ 9.5 \\
 & HSK4 & 66.2 $\pm$ 9.3 \\
 & HSK5 & 68.8 $\pm$ 10.2 \\
 & HSK6 & 75.5 $\pm$ 9.4 \\
 & HSK7-9 & 79.7 $\pm$ 12.5 \\
\bottomrule
\end{tabular}
\caption{Vocabulary coverage score statistics for gemini-2.5-flash.}
\label{tab:detail_vocab_coverage_gemini-2.5-flash}
\end{table}

\begin{table}[htb]
\centering
\small
\begin{tabular}{llc}
\toprule
\textbf{Lang.} & \textbf{Level} & \textbf{Simplified} \\
& & (Avg. $\pm$ Std.) \\
\midrule
EN & A1 & 39.3 $\pm$ 13.0 \\
 & A2 & 52.6 $\pm$ 12.0 \\
 & B1 & 70.2 $\pm$ 10.1 \\
 & B2 & 79.8 $\pm$ 8.5 \\
 & C1 & 81.7 $\pm$ 7.5 \\
 & C2 & 84.9 $\pm$ 7.2 \\
\midrule
JA & N5 & 39.4 $\pm$ 10.3 \\
 & N4 & 46.1 $\pm$ 10.7 \\
 & N3 & 62.5 $\pm$ 11.0 \\
 & N2 & 62.8 $\pm$ 9.6 \\
 & N1 & 73.5 $\pm$ 10.2 \\
\midrule
KO & TOPIK1 & 36.1 $\pm$ 11.9 \\
 & TOPIK2 & 44.9 $\pm$ 11.0 \\
 & TOPIK3 & 55.4 $\pm$ 11.1 \\
 & TOPIK4 & 64.7 $\pm$ 10.6 \\
 & TOPIK5 & 69.0 $\pm$ 10.0 \\
 & TOPIK6 & 72.1 $\pm$ 11.2 \\
\midrule
ZH & HSK1 & 36.2 $\pm$ 11.8 \\
 & HSK2 & 46.2 $\pm$ 12.5 \\
 & HSK3 & 58.4 $\pm$ 10.6 \\
 & HSK4 & 61.8 $\pm$ 9.7 \\
 & HSK5 & 65.3 $\pm$ 9.4 \\
 & HSK6 & 71.7 $\pm$ 8.1 \\
 & HSK7-9 & 78.9 $\pm$ 9.8 \\
\bottomrule
\end{tabular}
\caption{Vocabulary coverage score statistics for claude-sonnet-4-5-20250929.}
\label{tab:detail_vocab_coverage_claude-sonnet-4-5-20250929}
\end{table}

\subsection{\framework\ Result Statistics}

Table~\ref{tab:eval_Qwen3-4B-Instruct-2507-trained}, \ref{tab:eval_pgv_Qwen3-4B-Instruct-2507-trained}, \ref{tab:eval_gemma-3-4b-it-trained}, and \ref{tab:eval_pgv_gemma-3-4b-it-trained} demonstrate the evaluation result of \framework\ on Qwen3-4B-Instruct-2507 and gemma-3-4b-it with the Wikipedia and PGV test datasets.

\begin{table}[htb]
\centering
\small
\begin{tabular}{llccc}
\toprule
\textbf{Lang.} & \textbf{Level} & \textbf{Voc.} & \textbf{Sem.} & \textbf{Coh.} \\
\midrule
\multirow{7}{*}{EN} & A1 & 57.7 & 78.2 & 76.4 \\
 & A2 & 76.0 & 78.1 & 77.7 \\
 & B1 & 84.6 & 81.8 & 82.1 \\
 & B2 & 89.8 & 81.9 & 84.4 \\
 & C1 & 89.3 & 81.1 & 87.8 \\
 & C2 & 91.2 & 83.8 & 88.6 \\
\cmidrule(lr){2-5}
 & Total & 81.6 & 80.8 & 82.9 \\
\midrule
\multirow{6}{*}{JA} & N5 & 49.4 & 79.4 & 78.9 \\
 & N4 & 63.9 & 78.6 & 80.2 \\
 & N3 & 79.8 & 78.3 & 81.5 \\
 & N2 & 80.7 & 75.0 & 83.1 \\
 & N1 & 86.5 & 79.4 & 83.4 \\
\cmidrule(lr){2-5}
 & Total & 71.9 & 78.0 & 81.4 \\
\midrule
\multirow{7}{*}{KO} & TOPIK1 & 45.0 & 87.6 & 80.6 \\
 & TOPIK2 & 61.0 & 85.7 & 82.0 \\
 & TOPIK3 & 70.0 & 86.9 & 83.8 \\
 & TOPIK4 & 77.9 & 86.8 & 85.9 \\
 & TOPIK5 & 81.4 & 88.3 & 86.8 \\
 & TOPIK6 & 83.8 & 86.9 & 86.7 \\
\cmidrule(lr){2-5}
 & Total & 70.4 & 87.1 & 84.4 \\
\midrule
\multirow{8}{*}{ZH} & HSK1 & 60.0 & 78.0 & 80.5 \\
 & HSK2 & 71.6 & 76.3 & 81.3 \\
 & HSK3 & 79.4 & 76.4 & 82.9 \\
 & HSK4 & 82.7 & 76.3 & 84.5 \\
 & HSK5 & 84.1 & 77.2 & 84.9 \\
 & HSK6 & 89.4 & 77.2 & 85.0 \\
 & HSK7-9 & 92.9 & 75.1 & 86.4 \\
\cmidrule(lr){2-5}
 & Total & 80.2 & 76.6 & 83.7 \\
\bottomrule
\end{tabular}
\caption{Evaluation results of \framework\ on Qwen3-4B-Instruct-2507 with Wikipedia test data.}
\label{tab:eval_Qwen3-4B-Instruct-2507-trained}
\end{table}

\begin{table}[htb]
\centering
\small
\begin{tabular}{llccc}
\toprule
\textbf{Lang.} & \textbf{Level} & \textbf{Voc.} & \textbf{Sem.} & \textbf{Coh.} \\
\midrule
\multirow{7}{*}{EN} & A1 & 60.3 & 75.8 & 76.4 \\
 & A2 & 76.4 & 67.9 & 78.8 \\
 & B1 & 85.1 & 71.5 & 82.4 \\
 & B2 & 91.4 & 75.0 & 84.6 \\
 & C1 & 92.2 & 80.1 & 87.8 \\
 & C2 & 93.3 & 78.8 & 88.8 \\
\cmidrule(lr){2-5}
 & Total & 83.0 & 74.6 & 83.0 \\
\midrule
\multirow{6}{*}{JA} & N5 & 50.8 & 66.1 & 78.1 \\
 & N4 & 67.3 & 60.7 & 78.8 \\
 & N3 & 80.9 & 55.2 & 80.8 \\
 & N2 & 82.1 & 59.6 & 82.2 \\
 & N1 & 88.6 & 56.3 & 84.3 \\
\cmidrule(lr){2-5}
 & Total & 73.5 & 59.6 & 80.7 \\
\midrule
\multirow{7}{*}{KO} & TOPIK1 & 48.3 & 76.4 & 78.0 \\
 & TOPIK2 & 62.6 & 78.8 & 79.7 \\
 & TOPIK3 & 72.8 & 77.8 & 83.0 \\
 & TOPIK4 & 81.0 & 76.6 & 84.7 \\
 & TOPIK5 & 84.2 & 78.8 & 86.1 \\
 & TOPIK6 & 87.1 & 79.6 & 85.9 \\
\cmidrule(lr){2-5}
 & Total & 72.8 & 78.1 & 82.9 \\
\midrule
\multirow{8}{*}{ZH} & HSK1 & 56.2 & 63.5 & 81.5 \\
 & HSK2 & 73.0 & 68.2 & 82.1 \\
 & HSK3 & 79.6 & 69.6 & 84.1 \\
 & HSK4 & 83.1 & 61.7 & 85.5 \\
 & HSK5 & 87.2 & 72.7 & 87.0 \\
 & HSK6 & 89.3 & 69.1 & 85.7 \\
 & HSK7-9 & 92.7 & 64.5 & 87.5 \\
\cmidrule(lr){2-5}
 & Total & 80.8 & 67.1 & 84.8 \\
\bottomrule
\end{tabular}
\caption{Evaluation results of \framework\ on Qwen3-4B-Instruct-2507 with PGV test data.}
\label{tab:eval_pgv_Qwen3-4B-Instruct-2507-trained}
\end{table}

\begin{table}[htb]
\centering
\small
\begin{tabular}{llccc}
\toprule
\textbf{Lang.} & \textbf{Level} & \textbf{Voc.} & \textbf{Sem.} & \textbf{Coh.} \\
\midrule
\multirow{7}{*}{EN} & A1 & 57.7 & 76.4 & 76.2 \\
 & A2 & 74.4 & 78.4 & 77.6 \\
 & B1 & 84.7 & 81.8 & 79.7 \\
 & B2 & 89.2 & 80.7 & 82.3 \\
 & C1 & 90.1 & 80.1 & 84.0 \\
 & C2 & 92.0 & 81.9 & 84.6 \\
\cmidrule(lr){2-5}
 & Total & 81.5 & 79.9 & 80.8 \\
\midrule
\multirow{6}{*}{JA} & N5 & 51.9 & 77.2 & 77.2 \\
 & N4 & 64.5 & 76.1 & 79.1 \\
 & N3 & 79.4 & 77.8 & 79.6 \\
 & N2 & 80.8 & 73.9 & 80.7 \\
 & N1 & 87.0 & 76.6 & 80.9 \\
\cmidrule(lr){2-5}
 & Total & 72.5 & 76.3 & 79.5 \\
\midrule
\multirow{7}{*}{KO} & TOPIK1 & 49.0 & 85.0 & 79.0 \\
 & TOPIK2 & 63.4 & 81.8 & 81.3 \\
 & TOPIK3 & 73.1 & 84.1 & 81.7 \\
 & TOPIK4 & 79.2 & 84.0 & 82.3 \\
 & TOPIK5 & 82.7 & 86.0 & 83.4 \\
 & TOPIK6 & 84.6 & 85.9 & 83.7 \\
\cmidrule(lr){2-5}
 & Total & 72.5 & 84.5 & 82.0 \\
\midrule
\multirow{8}{*}{ZH} & HSK1 & 51.1 & 73.5 & 78.3 \\
 & HSK2 & 60.2 & 73.8 & 80.0 \\
 & HSK3 & 72.9 & 72.9 & 79.6 \\
 & HSK4 & 79.0 & 74.6 & 80.3 \\
 & HSK5 & 81.5 & 73.3 & 81.1 \\
 & HSK6 & 87.2 & 75.9 & 81.2 \\
 & HSK7-9 & 92.2 & 71.5 & 81.6 \\
\cmidrule(lr){2-5}
 & Total & 75.1 & 73.6 & 80.3 \\
\bottomrule
\end{tabular}
\caption{Evaluation results of \framework\ on Qwen3-4B-Instruct-2507 with Wikipedia test data.}
\label{tab:eval_gemma-3-4b-it-trained}
\end{table}

\begin{table}[htb]
\centering
\small
\begin{tabular}{llccc}
\toprule
\textbf{Lang.} & \textbf{Level} & \textbf{Voc.} & \textbf{Sem.} & \textbf{Coh.} \\
\midrule
\multirow{7}{*}{EN} & A1 & 58.7 & 76.1 & 77.2 \\
 & A2 & 75.6 & 67.6 & 78.0 \\
 & B1 & 85.8 & 67.7 & 79.9 \\
 & B2 & 91.0 & 74.4 & 82.4 \\
 & C1 & 92.4 & 75.4 & 84.4 \\
 & C2 & 92.9 & 71.2 & 85.0 \\
\cmidrule(lr){2-5}
 & Total & 82.7 & 72.1 & 81.1 \\
\midrule
\multirow{6}{*}{JA} & N5 & 51.5 & 67.1 & 77.2 \\
 & N4 & 66.5 & 64.4 & 78.5 \\
 & N3 & 80.9 & 60.7 & 79.2 \\
 & N2 & 81.4 & 64.3 & 80.1 \\
 & N1 & 88.8 & 65.1 & 81.5 \\
\cmidrule(lr){2-5}
 & Total & 73.4 & 64.3 & 79.2 \\
\midrule
\multirow{7}{*}{KO} & TOPIK1 & 54.1 & 76.5 & 78.1 \\
 & TOPIK2 & 64.8 & 81.7 & 80.1 \\
 & TOPIK3 & 73.6 & 73.0 & 81.4 \\
 & TOPIK4 & 82.6 & 74.5 & 82.0 \\
 & TOPIK5 & 85.7 & 78.0 & 83.2 \\
 & TOPIK6 & 87.8 & 76.9 & 83.2 \\
\cmidrule(lr){2-5}
 & Total & 74.9 & 76.6 & 81.4 \\
\midrule
\multirow{8}{*}{ZH} & HSK1 & 45.5 & 65.3 & 78.6 \\
 & HSK2 & 58.2 & 64.6 & 81.4 \\
 & HSK3 & 71.9 & 67.9 & 81.5 \\
 & HSK4 & 77.2 & 65.5 & 81.2 \\
 & HSK5 & 84.0 & 70.9 & 83.3 \\
 & HSK6 & 87.0 & 69.2 & 82.4 \\
 & HSK7-9 & 91.4 & 68.8 & 82.4 \\
\cmidrule(lr){2-5}
 & Total & 74.3 & 67.5 & 81.6 \\
\bottomrule
\end{tabular}
\caption{Evaluation results of \framework\ on Qwen3-4B-Instruct-2507 with Wikipedia test data.}
\label{tab:eval_pgv_gemma-3-4b-it-trained}
\end{table}

\begin{table*}[htb]
    \centering
    \small
    \begin{tabular}{lp{0.72\textwidth}}
    \toprule
    \textbf{Level} & \textbf{Simplified Text} \\
    \midrule
    Reference & Pigeon photography is an aerial photography technique invented in 1907 by the German apothecary Julius Neubronner, who also used pigeons to deliver medications. A homing pigeon was fitted with an aluminium breast harness to which a lightweight time-delayed miniature camera could be attached. Neubronner's German patent application was initially rejected, but was granted in December 1908 after he produced authenticated photographs taken by his pigeons. ...\\
    \midrule
    CEFR A1 & Pigeon photography is a way to take pictures from the air. It was made in 1907 by a man named Julius Neubronner from Germany. He also used pigeons to carry medicine. A kind of pigeon that can find its way home was given a small metal thing on its chest. On that chest, a small camera that did not weigh much and worked after a short time could be put. Neubronner wrote a plan in Germany, but first it was not accepted. In December 1908, it was allowed after he showed real pictures taken by his pigeons. ... \\
    \midrule
    CEFR A2 & Pigeon photography is a way of taking photos from the air. It was made in 1907 by a man named Julius Neubronner from Germany. He also used pigeons to carry medicine. A kind of pigeon that can find its way home was given a small metal thing on its chest. On that chest, a small camera that did not weigh much and worked after a short time could be put. Neubronner wrote a patent in Germany, but first it was not accepted. In December 1908, it was allowed after he showed real photos taken by his pigeons. ... \\
    \midrule
    CEFR B1 & Pigeon photography is a way of taking photos from the air that was first made in 1907 by a German man named Julius Neubronner. He also used pigeons to carry medicine. A type of pigeon that can find its way home was given a small metal piece on its chest, where a small camera that worked slowly and weighed very little could be put. Neubronner wrote a patent in Germany, but it was not accepted at first. In December 1908, it was finally allowed after he showed real photos taken by his pigeons. ... \\
    \midrule
    CEFR B2 & Pigeon photography is a way of taking photos from the air that was first created in 1907 by a German man named Julius Neubronner. He also used pigeons to carry medicine. A type of pigeon that can find its way home was given a small metal piece on its chest, where a very light camera — which worked after a short delay — could be placed. Neubronner's idea for a German patent was first turned down, but in December 1908, it was finally accepted after he showed real photos taken by his pigeons. ... \\
    \midrule
    CEFR C1 & Pigeon photography is a type of aerial photography that was first created in 1907 by a German man named Julius Neubronner, who also used pigeons to carry medicine. A pigeon capable of finding its way home was given an aluminium device on its chest, where a small, light camera with a delay feature could be placed. Neubronner's patent in Germany was first turned down, but later approved in December 1908 after he showed real photos taken by his pigeons. ... \\
    \midrule
    CEFR C2 & Pigeon photography is a method used in aerial imaging that was first created in 1907 by a German man named Julius Neubronner, who also used pigeons to carry medicines. A type of homing pigeon was equipped with an aluminium device on its chest, allowing a small, light camera with a delay feature to be attached. Neubronner's patent request in Germany was first turned down, but it was finally approved in December 1908 after he showed real photos taken by his pigeons. ... \\
    \bottomrule
    \end{tabular}
\caption{\framework\ generation examples by proficiency levels in English. The trained policy model is Qwen3-4B-Instruct-2507.}
\label{tab:example_generation_english}
\end{table*}

\begin{table*}[htb]
    \centering
    \small
    \begin{tabular}{lp{0.72\textwidth}}
    \toprule
    \textbf{Level} & \textbf{Simplified Text} \\
    \midrule
    Reference & ... \begin{CJK}{UTF8}{mj}집회와 결사의 자유, 언론의 자유 등의 자유권의 보장과 독일 의회의 수립, 그리고 헌법의 제정을 요구했다. 비록 독일에서의 혁명은 뚜렷한 족적을 남기지는 않았지만, 프리드리히가 어릴 때 목도한 이 자유주의는 훗날 그의 일생에 걸쳐 큰 영향력을 발휘하게 된다.\end{CJK}\\
    & (\textit{Translation: ... They demanded the guarantee of civil liberties such as freedom of assembly and association, freedom of the press, the establishment of a German parliament, and the enactment of a constitution. Although the revolution in Germany did not leave a distinct mark, the liberalism Friedrich witnessed in his youth would exert a profound influence throughout his life.'')}\\
    \midrule
    TOPIK1 & ... \begin{CJK}{UTF8}{mj}사람들은 모임과 함께 행동할 수 있고, 언론도 자유로이 쓸 수 있도록 하는 등의 자유를 받고, 독일의 회의를 만들고, 새로운 법을 만드는 것을 요구했다. 독일에서 일어난 혁명은 특히 큰 변화를 만들지는 않았지만, 프리드리히가 어릴 때 보았던 자유주의는 그 후 그의 삶 전체에서 큰 도움이 되었다.\end{CJK} \\
    \midrule
    TOPIK2 & ... \begin{CJK}{UTF8}{mj}사람들이 모여 이야기할 수 있고, 언론도 자유롭게 쓸 수 있도록 하는 등의 자유를 받고, 독일의 회의를 만들고, 새로운 법을 만드는 것을 요구했다. 독일에서 일어난 혁명은 특히 큰 변화를 가져오지는 않았지만, 프리드리히가 어릴 때 보았던 자유주의의 생각은 그 후 그의 삶 전체에서 큰 영향을 주게 되었다.\end{CJK} \\
    \midrule
    TOPIK3 & ... \begin{CJK}{UTF8}{mj}회의나 함께 행동할 수 있는 자유, 언론에서 자유롭게 말할 수 있는 권리 같은 중요한 권리들을 받고, 독일의 의회를 만들고, 새로운 헌법도 만들 것을 요구했다. 독일 내에서 일어난 혁명은 특히 큰 변화를 가져오지는 않았지만, 프리드리히가 어릴 때 보았던 자유주의의 생각은 그 후 그의 삶 전체에서 매우 큰 영향을 주게 되었다.\end{CJK} \\
    \midrule
    TOPIK4 & ... \begin{CJK}{UTF8}{mj}회의나 결사, 언론 등 자유를 보장하고, 독일의 의회를 만들고, 헌법을 만들어야 한다고 요구했다. 독일 내에서 일어난 혁명은 특히 큰 변화를 가져오지는 않았지만, 프리드리히가 어릴 때 보았던 자유주의의 모습은 그 후 그의 삶 전체에서 매우 중요한 영향을 주게 되었다.\end{CJK} \\
    \midrule
    TOPIK5 & ... \begin{CJK}{UTF8}{mj}회의나 단체 활동, 언론 사용 등 자유를 보장하고, 독일의 의회를 만들고 새로운 헌법을 만들어야 한다고 요구했다. 독일 내에서 일어난 혁명은 특히 명확한 결과를 가져오지는 않았지만, 프리드리히가 어릴 때 직접 보았던 자유주의의 개념은 그 후 그의 삶 전체에서 매우 중요한 영향을 주게 되었다.\end{CJK} \\
    \midrule
    TOPIK6 & ... \begin{CJK}{UTF8}{mj}회의나 단체 활동, 언론 사용 등 자유를 보장받을 수 있는 조건과 독일의 의회를 만들고, 헌법을 만들어야 한다는 요구를 제시했다. 독일 내에서 일어난 혁명은 특히 명확한 결과를 가져오지는 못했지만, 프리드리히가 어릴 때 직접 보았던 자유주의의 사례는 그 후 그의 삶 전체에서 매우 큰 영향을 주게 되었다.\end{CJK} \\
    \bottomrule
    \end{tabular}
\caption{\framework\ generation examples by proficiency levels in a non-English language (Korean). The trained policy model is Qwen3-4B-Instruct-2507.}
\label{tab:example_generation_korean}
\end{table*}

\section{Prompt}
\label{sec:prompts}

Table~\ref{tab:prompt_coherence} shows the exact prompt for coherence scoring via an evaluator LLM.

\section{Code and Data Release}

We release the \framework\ training code and available vocabulary level datasets online to contribute to the research community.\footnote{\texttt{https://github.com/jjhsnail0822/text-simplification}}

\section{Use of AI Assistants}

In this work, we use AI assistants for coding and checking grammatical errors.

\begin{table*}[htb]
\centering
\small
    \begin{tabular}{p{0.9\textwidth}}
        \toprule
You are evaluating \verb|{language}| text quality for a text simplification system.\\
\\
Given [ORIGINAL\_TEXT] and [SIMPLIFIED\_TEXT], focus ONLY on how natural and fluent the [SIMPLIFIED\_TEXT] reads as a rewrite of the [ORIGINAL\_TEXT]. Rate the NATURALNESS of the [SIMPLIFIED\_TEXT] as if it were written by a native speaker, strictly according to the following rules:\\
\\
100 = indistinguishable from a native human-written well-edited text\\
80-99 = highly natural with only minor unnatural phrasing\\
60-79 = generally understandable but contains multiple awkward and unnatural expressions\\
30-59 = sounds clearly machine-generated, frequently unnatural or repetitive\\
0-29 = extremely incoherent or clearly broken language\\
\\
Critical penalties:\\
- Strongly penalize repetitive template phrasing (e.g., repeating the same word/phrase many times to fill text).\\
- Strongly penalize awkward connective phrases or unnatural sentence patterns.\\
- Do NOT reward being 'simple' if it becomes unnatural. Simple but fully natural text should still receive a high score.\\
\\
Use the full 0–100 range. Reflect even small differences in naturalness with 1-point precision.\\
Output only a single integer from 0 to 100, and say nothing else.\\
\\
\textnormal{[}ORIGINAL\_TEXT]\\
\verb|{original_text}|\\
\\
\textnormal{[}SIMPLIFIED\_TEXT]\\
\verb|{simplified_text}|\\
        \bottomrule
    \end{tabular}
    \caption{Prompt for coherence scoring.}
\label{tab:prompt_coherence}
\end{table*}

\end{document}